\documentclass[11pt]{article}

\usepackage[preprint]{acl}
\usepackage{times}
\usepackage{latexsym}
\usepackage[T1]{fontenc}
\usepackage[utf8]{inputenc}
\usepackage{microtype}
\usepackage{inconsolata}
\usepackage{graphicx}
\usepackage{booktabs}
\usepackage{amsmath}
\usepackage{amssymb}
\usepackage{algorithm}
\usepackage{algpseudocode}

\providecommand{\storyblockstart}{%
  \par\begingroup
  \small
  \setlength{\parindent}{0pt}%
  \setlength{\parskip}{0.25em}%
}

\providecommand{\storyblockend}{%
  \par\endgroup
}

\providecommand{\storyheading}[1]{%
  \par\noindent\textbf{#1}\par
}

\providecommand{\storypara}[1]{%
  \par\noindent #1\par
}

\usepackage{booktabs}
\usepackage{array}
\usepackage{capt-of}    
\usepackage{needspace}  
\usepackage{placeins}   

\usepackage[most]{tcolorbox}
\usepackage{xcolor}

\definecolor{StorySparkGreen}{HTML}{6AD1A3}
\definecolor{StorySparkBg}{RGB}{248,248,248}

\newtcolorbox{promptbox}[2][]{%
  enhanced,
  breakable,
  colback=StorySparkBg,
  colframe=StorySparkGreen,
  colbacktitle=StorySparkGreen,
  coltitle=white,
  fonttitle=\bfseries\small,
  title={#2},
  title after break={#2 \textnormal{(continued)}},
  boxrule=0.8pt,
  arc=2mm,
  left=2mm,
  right=2mm,
  top=1.5mm,
  bottom=1.5mm,
  toptitle=1mm,
  bottomtitle=1mm,
  before skip=0.75\baselineskip,
  after skip=0.75\baselineskip,
  before upper={%
    \scriptsize
    \setlength{\parindent}{0pt}%
    \setlength{\parskip}{0.35em}%
  },
  #1
}

\newcommand{\slot}[1]{\textsc{#1}}

\title{StorySpark: Module-wise Evolutionary Search\\for Story Premise Generation}

\author{
{\bfseries Yang Yang$^{1,\dagger}$, Zining Zhong$^{1,\dagger}$, Qian Cao$^{2}$, Jindong Li$^{1}$,}\\
{\bfseries Boyun Xu$^{3}$, Kaishen Yuan$^{1}$, Menglin Yang$^{1}$, Yutao Yue$^{1,4,*}$}\\[0.35em]
{\normalfont\small $^{1}$The Hong Kong University of Science and Technology (Guangzhou), Guangzhou 511400, China}\\
{\normalfont\small $^{2}$Renmin University of China, Beijing, China}\\
{\normalfont\small $^{3}$Shandong University, Weihai, China}\\
{\normalfont\small $^{4}$Institute of Deep Perception Technology, JITRI, Wuxi 214000, China}\\[0.15em]
{\normalfont\small $^{\dagger}$Equal contribution. $^{*}$Corresponding author.}
}

\begin{document}
\maketitle

\begin{abstract}
A story premise is the creative spark from which a full narrative can grow. Yet LLM-based story generation has mostly emphasized later-stage planning, controllability, coherence, and prose expansion, while premise-level ideation remains comparatively underexplored. 
We introduce StorySpark, a module-wise evolutionary search framework for story premise generation. StorySpark operates over interpretable narrative modules such as background, persona, event, ending, and twist, treating each active module not as a static field to fill once, but as a local search space conditioned on the partial premise built so far. For each module, it generates alternatives, evaluates them in context, refines them through feedback-driven mutation and recombination, preserves complementary strengths with Pareto-guided selection, and reallocates frontier capacity to balance branch coverage with promising directions.
Multi-view automatic and human evaluations show that StorySpark produces stronger final premises than competitive baselines, with especially consistent gains in originality; when expanded with the same story writer, its premises also lead to higher-quality downstream stories while maintaining completeness, fascination, and diverse usable narrative directions.
\end{abstract}

\section{Introduction}

\begin{quote}
\small
\emph{``That premise will suggest the essence of the story, and we will use that to figure out how to develop it so as to get the most out of the idea.''}

\hfill --- John Truby, \textit{The Anatomy of Story}
\end{quote}

\noindent High-quality stories matter across fiction and screenwriting \citep{field2005screenplay,truby2008anatomy,cron2012wired}, as well as games, collaborative storytelling, and human-AI narrative systems \citep{akoury2020storium,mirowski2023co}. 
A successful story requires fluent prose, coherent progression, and emotional engagement, but these qualities are difficult to sustain without a compelling story premise \citep{field2005screenplay,lyons2015anatomy,truby2008anatomy}. 
Such a premise is a compact early seed that captures a story's main idea, foundation, and trajectory, and supports later expansion into narrative text \citep{lyons2015anatomy,ma2024mops}.

\begin{figure}[!t]
  \centering
  \includegraphics[width=\columnwidth]{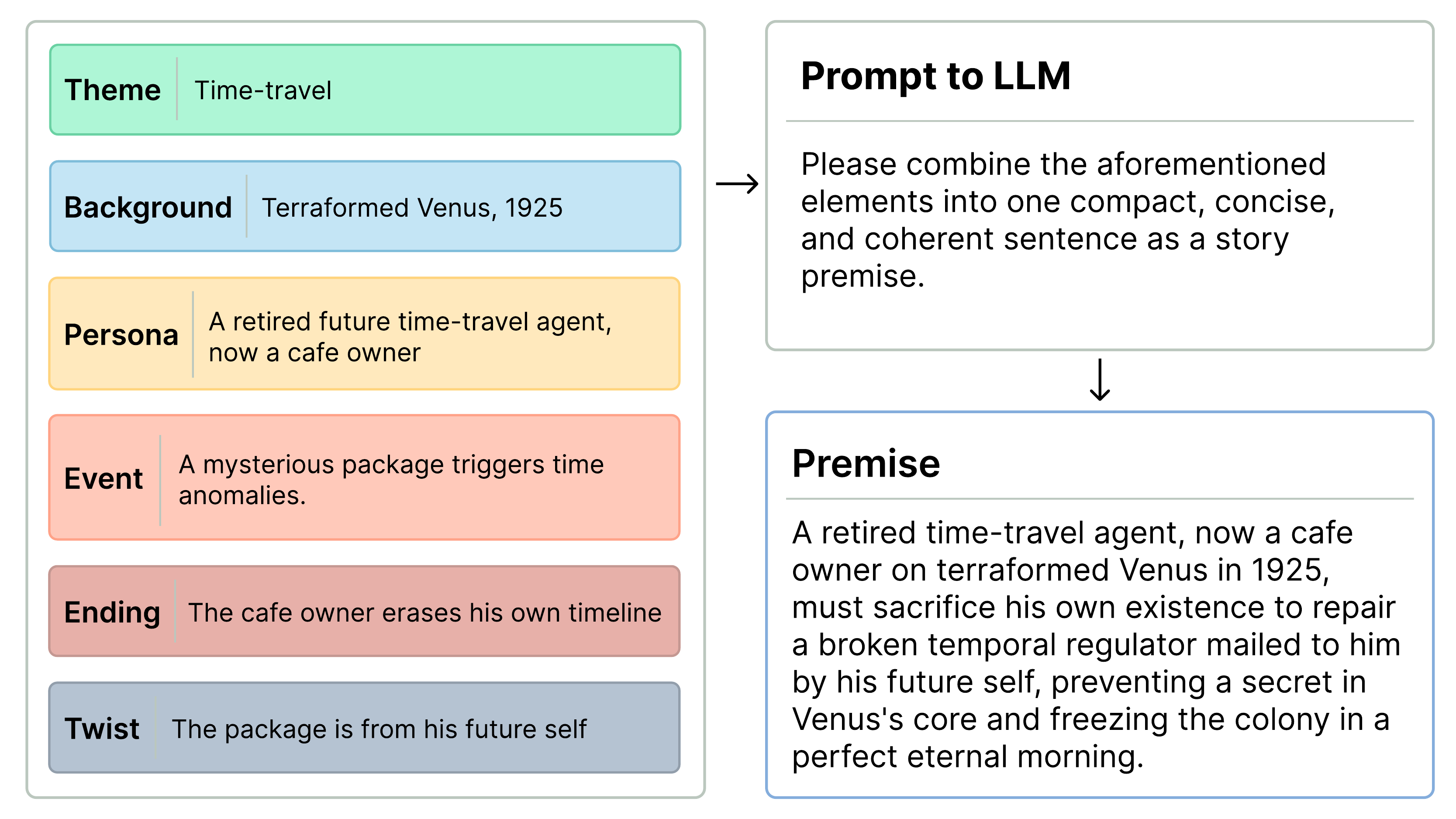}
  \vspace{-0.5em}
  \caption{A premise as the creative spark between an input theme and later outline or story expansion.}
  \label{fig:premise-modules}
  \vspace{-0.4em}
\end{figure}

Recent LLM-based story generation has advanced later-stage narrative development through hierarchical and plan-and-write generation \citep{fan2018hierarchical,yao2019plan}, controllable plot and event modeling \citep{martin2018event,tambwekar2018controllable}, long-form coherence, outline control, and recursive revision \citep{park2024longstory,yang2022re3,yang2023doc,zhu2023end,xia2025storywriter}, and interactive or machine-in-the-loop authoring \citep{akoury2020storium,mirowski2023co}. 
However, most of these efforts focus on how an existing narrative seed is planned, controlled, revised, or expanded into longer text. 
The earlier ideation problem---how to produce a premise that is worth developing in the first place, as illustrated in Figure~\ref{fig:premise-modules}---has received comparatively less attention.

DOC improves long-story coherence through detailed outline control, but its outline-driven pipeline does not specifically optimize the premise as an early-stage design artifact \citep{yang2023doc}. 
DPWriter introduces planning-level branching to promote diversity in complete creative-writing responses, but its search is aimed at diverse downstream generations rather than quality-guided premise construction \citep{cao2026dpwriter}. 
MoPS is more premise-centered: it represents a premise as modular components, evaluates premise quality, and synthesizes sampled module paths from a nested inventory into complete premises \citep{ma2024mops}. 
However, its modular candidates are largely fixed after induction and are mainly used through path sampling and final synthesis, without repeatedly scoring, revising, recombining, or reallocating active candidates while the partial premise is being built. 
Weak local choices may therefore persist, and a fixed candidate structure may fail to adapt capacity across directions with different promise or different trade-offs among completeness, fascination, and originality.

In this paper, we introduce StorySpark, a module-wise evolutionary search framework for story premise generation. 
StorySpark keeps the modular premise space introduced by MoPS, but turns construction into quality-guided search over a growing partial premise. 
Under a fixed parent prefix, it evolves the active module through feedback-driven mutation and recombination, allowing weak initial candidates to be improved before their branches are discarded. 
It then uses Pareto-guided local selection to preserve candidates with different strengths across completeness, fascination, and originality, and reserve-wildcard frontier allocation to balance branch coverage with extra capacity for promising directions \citep{deb2000fast,  mouret2015illuminating,pugh2016quality}. 
Across premise-level scoring, ablation and process analysis, pairwise validation, story-level transfer, and quality-aware diversity analysis, StorySpark consistently improves premise quality while showing that the gain is not merely an artifact of a single pointwise judge.


In summary, our contributions are threefold: 
\textbf{(1) Evolutionary premise search.} To the best of our knowledge, we are the first to formulate story premise generation as evolutionary search, moving premise-level ideation beyond one-shot prompting and static modular sampling. 
\textbf{(2) Prefix-conditioned evolution.} We introduce a module-wise framework with partial-premise scoring, feedback-driven mutation and recombination, Pareto-guided local selection, and reserve-wildcard frontier allocation. 
\textbf{(3) Premise-and-story validation.} We show through automatic and human scoring, ablations, pairwise validation, story-level transfer, and quality-aware diversity analysis that StorySpark improves premise quality, especially originality, without relying on a single evaluation view.
\section{Related Work}

\subsection{Story Generation and Premise Design}

\textbf{Narrative planning, expansion, and premise}
Automatic story generation has moved from short-prompt continuation toward structured pipelines for later-stage narrative development. 
Prior work studies hierarchical and plan-and-write generation \citep{fan2018hierarchical,yao2019plan}, controllable plot and event modeling \citep{tambwekar2018controllable,martin2018event,fan2019strategies}, long-form coherence, outline control, and revision \citep{park2024longstory,yang2022re3,yang2023doc,zhu2023end,xia2025storywriter}, and diverse planning-level branching for creative writing \citep{cao2026dpwriter}. 
These methods improve how systems develop or diversify narrative material after an instruction, seed, outline, or intent is provided, but they do not directly optimize the compact premise seed itself. 
Premise-centered work takes a more direct step: MoPS represents a premise as a modular object composed of theme, background, persona, and plot-related components, then samples module paths from a nested inventory and synthesizes them into complete premises \citep{ma2024mops}. 
StorySpark adopts this modular premise space, but changes the optimization regime: rather than sampling final designs from a static inventory, it searches over each module under a fixed partial-premise prefix, allowing candidate modules to be scored, revised, recombined, and selected while the premise is being built.

\textbf{Interactive creative ideation.}
Interactive writing and story-ideation systems show that generated suggestions are most useful when they fit an evolving creative context and help writers explore alternatives rather than commit to a single continuation \citep{lee2022coauthor,akoury2020storium}. 
TaleBrush\citep{chung2022talebrush} gives writers sketch-based control for story ideation, while ABScribe\citep{reza2024abscribe} supports rapid exploration and comparison of multiple LLM-generated writing variations \citep{chung2022talebrush,reza2024abscribe}. 
These systems highlight a premise-level challenge: early ideation often requires repeatedly exploring, comparing, and revising alternative narrative seeds before committing to a full story. 
StorySpark systematizes this exploration process by formulating premise construction as module-wise evolutionary search, using automatic scoring, mutation, recombination, and selection to produce premise candidates that can be inspected, compared, and further developed before later story expansion.

\subsection{LLM-driven Search and Evaluation for Creative Generation}

\textbf{LLM-driven evolutionary and iterative search.}
Recent LLM-driven search and refinement systems suggest that open-ended artifacts can be improved through repeated proposal, evaluation, selection, and feedback-driven revision. In code and algorithm discovery, FunSearch and AlphaEvolve place language models inside search loops with automatic evaluators, while ShinkaEvolve studies more sample-efficient open-ended program evolution \citep{romera2024mathematical,novikov2025alphaevolve,lange2025shinkaevolve}. In language generation, Self-Refine shows that feedback-driven revision can improve model outputs through iterative self-feedback \citep{madaan2023self}. StorySpark draws on both views, but moves the search loop to a different granularity: instead of optimizing complete programs or revising complete generated outputs, it evolves current-module candidates under fixed premise prefixes. This smaller and more controlled search unit makes candidates easier to compare and links score changes more directly to specific narrative decisions such as background, persona, event, ending, or twist.

\textbf{Evaluation for creative search.}
Evaluation is tightly coupled to this search design. MoPS evaluates premises with pointwise LLM scores for fascination, completeness, and originality, and measures diversity through breadth and density \citep{ma2024mops}. More broadly, recent evaluator work studies rubric-based LLM judging, LLM-as-judge protocols, and pairwise ranking \citep{kim2024prometheus,zheng2023judging,jiang2023llm,cao2025evaluating}, while classical pairwise models provide foundations for relative preference estimation \citep{bradley1952rank,glickman1999rating}. Pointwise scores are efficient enough to guide repeated local search, but creative artifacts are difficult to validate with a single calibrated scalar judgment. StorySpark therefore uses pointwise judges as search-time signals and validates final premises through complementary pointwise, pairwise, downstream, and diversity-oriented views.

\section{Method}
\label{sec:method}

\begin{figure*}[t]
  \centering
  \includegraphics[width=\textwidth]{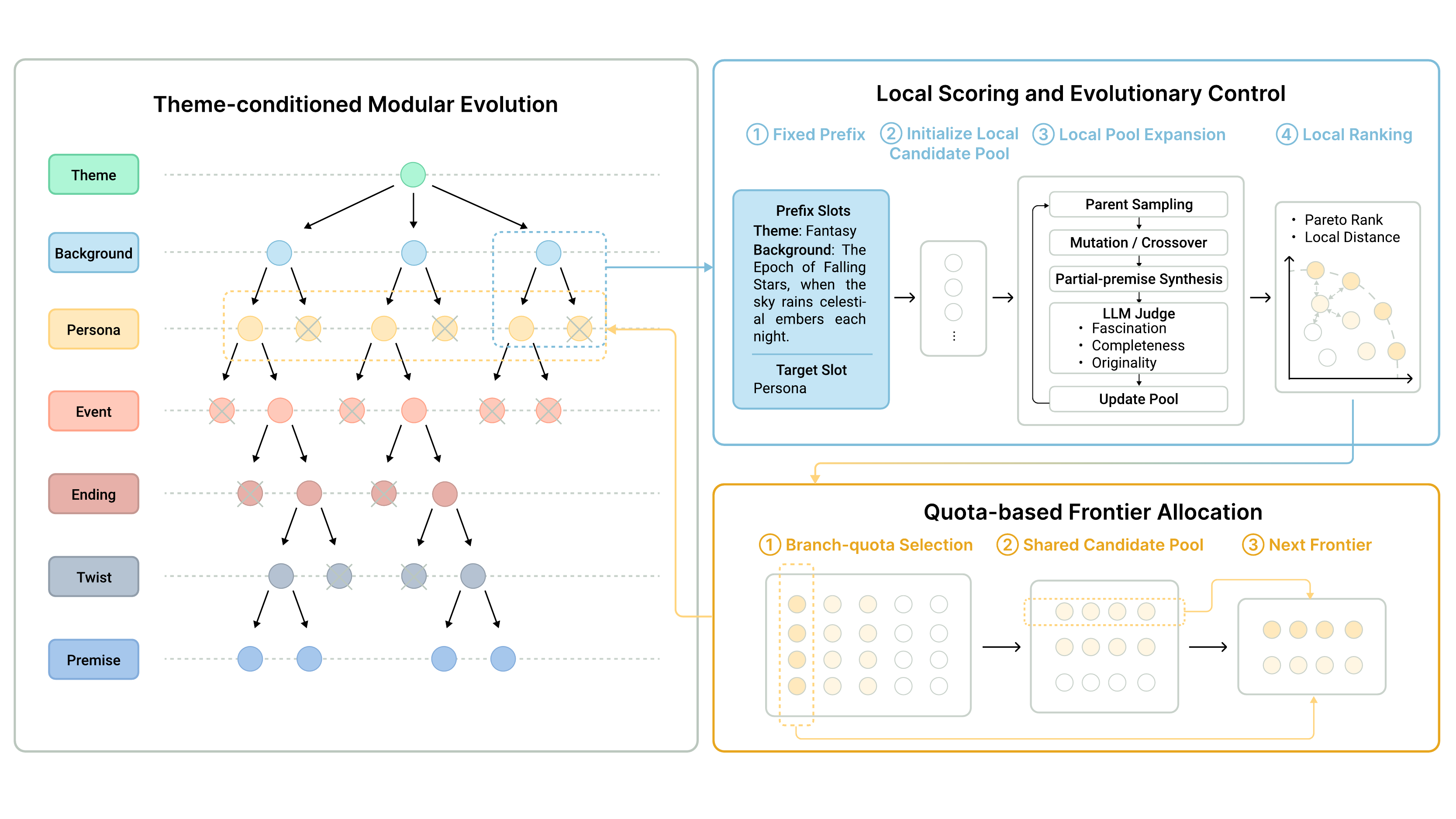}
  \vspace{-3.5em}
    \caption{Overview of StorySpark. Given a theme, each frontier branch evolves current-module candidates under a fixed prefix using contextual scoring; quota-based allocation promotes selected descendants to the next frontier.}
  \label{fig:storyspark_algorithm}
  \vspace{-0.8em}
\end{figure*}

StorySpark is a module-wise evolutionary search framework for story premise generation, as illustrated in Fig~\ref{fig:storyspark_algorithm}. It follows the modular premise view of MoPS~\citep{ma2024mops}, but changes how this space is used. Instead of sampling a complete key path from a static module inventory, StorySpark grows a premise one module at a time through quality-guided search. Given a fixed \slot{theme}, it maintains a frontier of partial module paths. For each frontier branch, the current module is searched under the branch's fixed prefix: candidates are generated, placed back into context for scoring, improved through mutation and crossover, and promoted through quota-based allocation. This keeps the modular structure interpretable while making each module revisable during construction.

We use \(\alpha\) for the \slot{theme}, which serves as the root condition rather than an active module. 
The \emph{active module sequence} is \(\mathcal{M}=(m_1,\ldots,m_5)\), corresponding to \slot{background}, \slot{persona}, \slot{event}, \slot{ending}, and \slot{twist}. 
A \emph{module candidate} \(c_t\) fills the current module \(m_t\); a \emph{module prefix} \(x_{1:t-1}\) is the upstream path selected so far; a \emph{materialized partial premise} \(z_t\) is the temporary natural-language premise synthesized from the theme, prefix, and current candidate for scoring; and a \emph{frontier} \(\mathcal{F}^{\alpha}_t\) is the set of partial module paths retained after module \(m_t\).

\subsection{Theme-conditioned Modular Evolution}
\label{sec:modular_evolution}

StorySpark represents a premise as an ordered module path conditioned on a fixed theme. Following the MoPS-style modular view, \slot{background} specifies the story world, \slot{persona} introduces the driving character configuration, and \slot{event}, \slot{ending}, and \slot{twist} specify the plot trajectory and payoff. We treat these five components as active modules and search them under an increasingly specific narrative prefix.

For each theme \(\alpha\), StorySpark initializes \(\mathcal{F}_0^\alpha=\{\emptyset\}\). At module \(m_t\), each node from the previous frontier
\begin{equation}
  x_{1:t-1}^{(b)}
  =
  (c_1^{(b)},\ldots,c_{t-1}^{(b)})
  \in
  \mathcal{F}_{t-1}^{\alpha},
\end{equation}
serves as a parent branch \(b\), where \(c_i^{(b)}\) is the selected candidate for module \(m_i\). StorySpark then opens a local candidate pool \(\mathcal{P}_t^{(b)}\) for the current module under this fixed prefix. For example, when searching for a persona, the LLM observes the theme and selected background, but proposes only persona candidates. Thus, generation remains local while scoring remains prefix-conditioned.

At module \(m_t\), \(B_t\) denotes the per-branch frontier capacity, \(R_t\le B_t\) the reserved quota, and \(G_t\) the target local-pool size after evolution. For each parent branch, the initializer proposes \(B_t\) seed candidates conditioned on the theme and prefix. Local evolution expands the pool to
\begin{equation}
  \mathcal{P}_t^{(b)}
  =
  \{c_t^{(b,1)},\ldots,c_t^{(b,G_t)}\}.
\end{equation}
Each \(c_t^{(b,i)}\) is a candidate for the current module rather than a complete premise.

StorySpark expands each local pool with mutation and crossover. Mutation rewrites one candidate using its previous text and judge feedback, while crossover fuses two candidates from the same pool using feedback from both parents. Both operators keep the theme and upstream modules fixed, and only modify the current module. Typed modules such as background and persona are checked against predefined module-type constraints.

\subsection{Local Scoring and Evolutionary Control}
\label{sec:local_control}

A module fragment cannot be judged reliably in isolation, because its value depends on the prefix it extends. StorySpark therefore first places each current-module candidate back into context. Let \(\operatorname{Mat}_t(\cdot)\) denote the partial-premise synthesis operation. Given the theme, fixed prefix, current module identity, and current-module candidate, it produces a contextually readable partial premise:
\begin{equation}
  z_t^{(b,i)}
  =
  \operatorname{Mat}_t
  \left(
    \alpha,\,
    x_{1:t-1}^{(b)},\,
    m_t,\,
    c_t^{(b,i)}
  \right).
\end{equation}
\(z_t^{(b,i)}\) is a temporary premise used to score \(c_t^{(b,i)}\) under branch \(b\).

An LLM judge scores each materialized partial premise along three quality dimensions:
\begin{equation}
  \mathbf{q}(z_t^{(b,i)})
  =
  (q_{\mathrm{fas}}, q_{\mathrm{com}}, q_{\mathrm{ori}}),
\end{equation}
where the dimensions denote fascination, completeness, and originality. For intermediate states, completeness is interpreted relative to the available prefix: it measures whether the partial premise provides coherent and useful information for subsequent module development, rather than whether it already forms a complete final premise. For brevity, we write \(\mathbf{q}(c_t^{(b,i)})\) for the score assigned through its materialized partial premise.

The three dimensions are not collapsed into a single scalar during local selection. A candidate may be original but incomplete, or complete but less surprising; averaging too early can remove alternatives that later modules could make useful. StorySpark therefore treats the scores as a multi-objective vector. Candidate \(a\) Pareto-dominates candidate \(b\) if it is no worse on all dimensions and strictly better on at least one. Non-dominated sorting assigns each candidate a Pareto rank \(r(c)\), where a smaller rank is better.

To avoid filling a local pool with near-duplicates, StorySpark also computes local crowding from embeddings of current-module candidate texts. Since all candidates in one pool share the same theme and prefix, this directly measures non-redundancy among alternative fillings of the current module.

Parents for further variation are sampled using both quality and diversity:
\begin{equation}
\begin{aligned}
w(c)
&=
\frac{\epsilon+\tilde{d}(c)}{\max(r(c),1)},\\
p(c)
&=
\frac{w(c)}
{\sum_{c'\in\mathcal{P}_t^{(b)}} w(c')}.
\end{aligned}
\end{equation}
where \(\tilde{d}(c)\) is normalized local crowding and \(\epsilon\) is a smoothing constant. At each expansion step, StorySpark applies mutation with probability \(\mu_t\) and crossover otherwise, until the local pool reaches \(G_t\). Invalid, duplicate, type-conflicting, or materialization-failed candidates are filtered before entering the local pool.

\subsection{Quota-based Frontier Allocation}
\label{sec:frontier_allocation}

After local evolution, StorySpark allocates the next frontier across competing parent branches. A purely global top-\(k\) rule may collapse the search into a few strong branches too early, while a purely branch-local rule may preserve coverage but weaken cross-branch competition. StorySpark therefore uses quota-based frontier allocation to balance branch coverage with competition among promising candidates.

The allocation has two parts. The \emph{reserved quota} gives each parent branch \(R_t\) positions, filled by strong and diverse candidates from its own local pool. The \emph{shared quota} sends runner-up candidates from all branches into a shared pool, where they compete for the remaining capacity at the same module layer. Thus, every branch keeps minimum coverage, while branches with multiple strong candidates can receive extra positions.

Reserved-quota and shared-quota candidates are selected by the following sorting keys:
\begin{equation}
\vcenter{\hbox{\small$\displaystyle
\begin{aligned}
\textsc{Reserved}:&\quad
(\mathrm{PR}\uparrow,\mathrm{LC}\downarrow,\mathrm{TB}\uparrow),\\
\textsc{Shared}:&\quad
(\mathrm{GPR}\uparrow,\mathrm{Ov}\downarrow,\mathrm{LC}\downarrow,\mathrm{TB}\uparrow),
\end{aligned}
$}}
\label{eq:quota_sort}
\end{equation}
where PR, GPR, LC, TB, and Ov denote local Pareto rank, global Pareto rank, local crowding, deterministic tie-breaking, and average quality score. Sorting uses ascending order for PR, GPR, and TB, and descending order for LC and Ov. GPR is recomputed over the shared pool, while LC is retained from each candidate's original local pool.

We choose \(G_t \ge B_t\) so that each local pool can supply its reserved-quota candidates while retaining runner-up candidates for shared-quota competition. 
The next frontier size and shared-quota capacity are
\begin{equation}
\begin{aligned}
\left|\mathcal{F}^{\alpha}_t\right|
&= B_t \left|\mathcal{F}^{\alpha}_{t-1}\right|,\\
H_t
&= (B_t - R_t)\left|\mathcal{F}^{\alpha}_{t-1}\right|.
\end{aligned}
\end{equation}
where \(H_t\) is the shared-quota capacity at module \(m_t\). Each promoted current-module candidate is attached back to its original parent prefix:
\begin{equation}
  x_{1:t}^{(b,i)}
  =
  (x_{1:t-1}^{(b)}, c_t^{(b,i)}).
\end{equation}
Thus, the shared quota introduces cross-branch competition but not cross-branch recombination.

After the twist module is completed, each full module path \((\alpha,c_1,\ldots,c_5)\) is synthesized by an LLM into a compact final premise and evaluated with the same three quality dimensions. Thus, final scores measure the completed premise rather than only intermediate partial states. Budget values, decoding settings, validity filtering, implementation details, and full pseudocode are provided in the experimental setup and Appendix~\ref{app:algorithm}. Prompt templates are provided in Appendix~\ref{app:prompts}.

\section{Experiment Setup}

We evaluate StorySpark using the MoPS genre domains, released baseline assets, and three shared quality dimensions. We assess final-premise quality, quality-conditioned coverage, module-wise search behavior, premise-to-story transfer, pairwise preference, independent rerating, and human ratings.

\subsection{Baselines and Comparison Sets}

We compare StorySpark with six baselines in three groups: modular synthesis, \textbf{Modular Story Premise Synthesis (MoPS)}~\citep{ma2024mops}; direct prompting, \textbf{Vanilla prompting (VIL)} and \textbf{Complex prompting (CPX)}; and external-source premise assets released with MoPS, \textbf{DOC}~\citep{zhu2023end}, \textbf{WritingPrompts (WP)}~\citep{fan2018hierarchical}, and \textbf{Storium (STM)}~\citep{akoury2020storium}.

\noindent\textbf{Modular baseline.}
MoPS samples paths from a modular premise inventory and synthesizes selected modules into complete story premises.

\noindent\textbf{Direct prompting.}
VIL directly prompts the generation backend to produce complete premises. CPX follows the Complex baseline in MoPS by conditioning direct prompting on a small set of example premises.

\noindent\textbf{External sources.}
DOC provides LLM-generated premise-like seeds from prior long-story generation work; WP provides Reddit writing-prompt titles; and STM provides structured game descriptions from a collaborative storytelling platform. StorySpark and MoPS use the same 14 MoPS genre domains, while DOC, WP, and STM are evaluated as aggregate premise sources.

\subsection{Evaluation Criteria and Judges}

\noindent\textbf{Pointwise quality.}
For automatic evaluation, final premises and expanded stories receive 0--100 scores for fascination, completeness, and originality; Overall is their arithmetic mean. The three dimensions capture narrative hook and emotional pull, sufficient story information for coherent development, and freshness beyond generic templates, respectively. Human ratings use the same dimensions for premises only, on a 1--5 Likert scale, and are reported separately from automatic scores.

\noindent\textbf{Quality-conditioned coverage.}
For semantic diversity, we follow the MoPS-style breadth view: final premises are embedded, projected into a two-dimensional t-SNE space, and measured by the covered area in this semantic plane. We compute breadth after filtering premises by automatic Overall thresholds and weight it by the retained fraction. This measures semantic coverage among usable high-quality premises rather than low-quality outliers.

\noindent\textbf{Pairwise preference.}
Pairwise evaluation provides a relative validation view that is less sensitive to absolute score calibration. An automatic judge compares anonymized premise pairs and returns A, B, or TIE. We summarize the outcomes with direct StorySpark-vs.-baseline preference scores and an Elo-style aggregate ranking.

\noindent\textbf{Model backends.}
DeepSeek-V4-Flash is used for generation, search-time feedback, primary pointwise scoring, pairwise judging, and story expansion. GPT-5.2 is used only as an independent evaluator: it rerates existing final-premise artifacts, expanded stories, and anonymized pairwise comparisons, without participating in generation or search. Generation calls use temperature 0.7, while evaluation calls are deterministic.

\subsection{Evaluation Experiments}
We summarize the main evaluation experiments below. Additional operational details, including artifact counts, artifact reuse, pairwise aggregation, story transfer, human evaluation, and quality-conditioned coverage, are provided in Appendix~\ref{app:experimental_setup}.

\noindent\textbf{Main premise comparison.}
We evaluate 1,000 final premises per method for automatic quality and quality-conditioned coverage.

\noindent\textbf{Mechanism and process analysis.}
We compare full StorySpark with variants that remove local evolution, dynamic competition, or both, evaluating 4,032 final premises per variant. We also score materialized partial premises at each module depth to analyze how quality and semantic breadth change during search.

\noindent\textbf{Premise-to-story transfer.}
We sample 100 premises per method and expand them with the same fixed StoryWriter-style pipeline. Since the writer is held constant, differences in story quality mainly reflect the premise seeds.

\noindent\textbf{Pairwise validation.}
We sample 100 premises per method from all seven methods and compare anonymized premise pairs. 

\noindent\textbf{Independent and human validation.}
The independent automatic judge rerates the main premise-quality artifacts. Human evaluation uses 20 anonymized premises per method from StorySpark, MoPS, Storium, and WritingPrompts, rated on Fascination, Completeness, and Originality.

\begin{table*}[t]
  \centering
  \footnotesize
  \setlength{\tabcolsep}{2.0pt}
  \begin{tabular}{@{}lccccrrrr@{}}
    \toprule
    \textbf{Method} & \multicolumn{4}{c}{\textbf{Premise Quality}} & \multicolumn{4}{c}{\textbf{Quality-conditioned Coverage}} \\
    \cmidrule(lr){2-5}\cmidrule(lr){6-9}
     & Fascination & Completeness & Originality & Overall
     & \(\tau=80\) & \(\tau=82\) & \(\tau=84\) & \(\tau=86\) \\
    \midrule
    StorySpark & \textbf{87.78} $\pm$ 1.40 & 87.45 $\pm$ 4.35 & \textbf{83.37} $\pm$ 5.00 & \textbf{86.20} $\pm$ 1.96 & \underline{6598} & \underline{6354} & \textbf{6044} & \textbf{3876} \\
    CPX~\citep{ma2024mops} & \underline{87.71} $\pm$ 1.28 & \textbf{88.30} $\pm$ 3.76 & \underline{78.96} $\pm$ 3.70 & \underline{84.99} $\pm$ 1.90 & 5810 & 5434 & 4431 & \underline{2160} \\
    MoPS~\citep{ma2024mops} & 87.09 $\pm$ 1.41 & \underline{87.83} $\pm$ 4.67 & 77.78 $\pm$ 3.75 & 84.23 $\pm$ 2.25 & \textbf{7127} & \textbf{6387} & \underline{4532} & 1863 \\
    VIL~\citep{ma2024mops} & 86.87 $\pm$ 1.60 & 85.42 $\pm$ 6.19 & 76.46 $\pm$ 4.29 & 82.92 $\pm$ 2.80 & 5293 & 4059 & 2275 & 490 \\
    DOC~\citep{zhu2023end} & 80.36 $\pm$ 10.44 & 77.40 $\pm$ 15.36 & 56.29 $\pm$ 19.01 & 71.35 $\pm$ 13.60 & 2095 & 820 & 26 & 0 \\
    STM~\citep{akoury2020storium} & 79.79 $\pm$ 13.77 & 72.48 $\pm$ 19.80 & 56.43 $\pm$ 25.33 & 69.57 $\pm$ 18.28 & 4281 & 2272 & 399 & 30 \\
    WP~\citep{fan2018hierarchical} & 82.26 $\pm$ 11.66 & 61.09 $\pm$ 20.38 & 56.40 $\pm$ 28.57 & 66.58 $\pm$ 18.13 & 1237 & 655 & 132 & 3 \\
    \bottomrule
  \end{tabular}
  \caption{Main premise quality and quality-conditioned coverage results under the DeepSeek-V4-Flash judge on 1,000 premises per method. Coverage is weighted t-SNE breadth after filtering by Overall cutoff \(\tau\); bold and underline denote the best and second-best results.}
  \vspace{-1em}
  \label{tab:premise_quality}
\end{table*}

\label{sec:experiment_setup}
\section{Results}
\label{sec:results}

\subsection{Premise-level Quality and High-quality Diversity}
\label{sec:premise_quality}

Table~\ref{tab:premise_quality} shows that StorySpark achieves the best premise-level Overall score, with the largest gain on Originality. Compared with the strongest baseline, StorySpark improves Overall by 1.21 points and Originality by 4.41 points, while remaining comparable on Fascination and Completeness. This indicates that the main advantage comes from finding fresher and less templated story seeds, rather than only improving structural completeness.

Quality-conditioned coverage further shows that StorySpark places more diverse premises in the high-quality region. MoPS has the largest coverage at lower cutoffs, but StorySpark becomes stronger as the quality threshold rises. At the highest cutoff, its coverage is about 1.8 times that of the strongest baseline. This suggests a shift in the quality--diversity frontier: more semantically distinct StorySpark premises remain after stringent quality filtering.

\begin{figure}[t]
  \centering
  \includegraphics[width=\columnwidth]{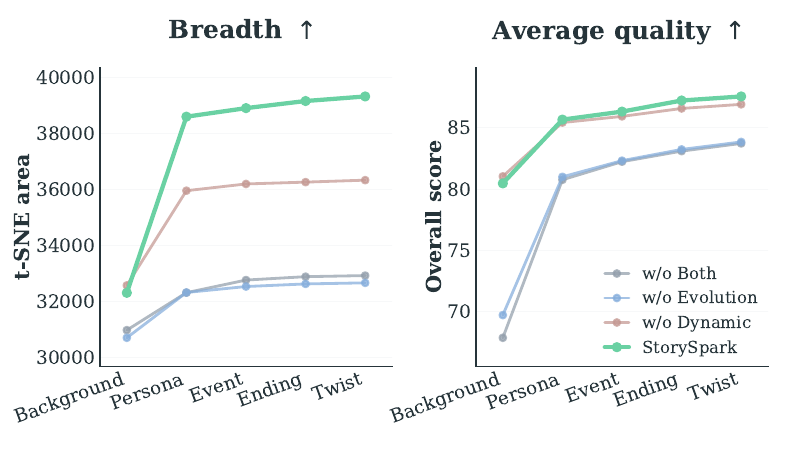}
  \vspace{-2em}
  \caption{module-level ablation process curves. Breadth denotes t-SNE area over retained partial premises, and quality denotes mean Overall.}
  \vspace{-1em}
  \label{fig:ablation_process}
\end{figure}

\subsection{Ablation and Process Analysis}
\label{sec:ablation_process}

Figure~\ref{fig:ablation_process} summarizes the ablation and module-level process curves. The full StorySpark system maintains the strongest quality trajectory and separates from the ablated variants from the persona stage onward, while its breadth does not collapse. Removing local evolution causes the largest quality drop, indicating that feedback-guided mutation and crossover are the main source of improved premise quality. Removing reserve--wildcard frontier allocation has a smaller standalone effect, but still weakens the trajectory, suggesting that frontier-level competition provides an additional selection benefit.

The two components are therefore best understood as complementary rather than equally strong in isolation. Local evolution creates better current-module candidates under a fixed prefix, whereas reserve--wildcard allocation decides how these improved candidates are carried forward across branches. The reserve quota preserves branch coverage, and the wildcard quota lets strong runner-up candidates compete across prefixes. This combination explains why the full model performs best: evolution enriches the local candidate pool, and frontier allocation converts those local gains into a stronger and more diverse next-step search frontier.

\begin{figure}[t]
  \centering
  \includegraphics[width=\columnwidth]{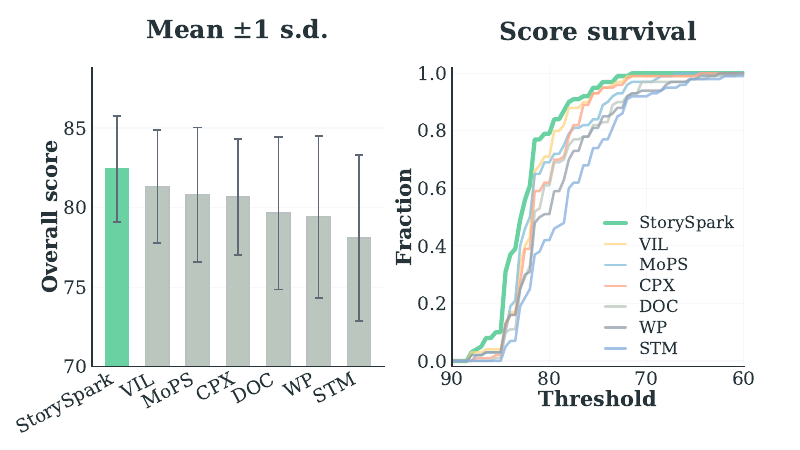}
  \vspace{-2em}
  \caption{Premise-to-story transfer. Left: mean story Overall with \(\pm 1\) s.d.; right: score-survival curve over Overall thresholds.}
  \vspace{-1em}
  \label{fig:story_transfer}
\end{figure}

\subsection{Premise-to-story Transfer}
\label{sec:story_transfer}

Figure~\ref{fig:story_transfer} shows that the premise-level gains transfer to downstream story generation. With the same story writer held fixed across methods, StorySpark ranks first on story-level Overall, indicating that its stronger premises remain useful after long-form expansion. The survival curve further shows that StorySpark has a larger high-quality tail: at Overall \(>84\), it retains 37\% of expanded stories, roughly twice the strongest baseline.

This suggests that StorySpark improves not only average story quality, but also the likelihood that a generated premise can support a high-scoring complete story. The result is consistent with the premise-level findings: more original and higher-quality story seeds lead to stronger downstream expansions rather than being washed out during generation.

\subsection{Independent Pairwise Validation}
\label{sec:pairwise_validation}

\begin{figure}[t]
  \centering
  \includegraphics[width=\columnwidth]{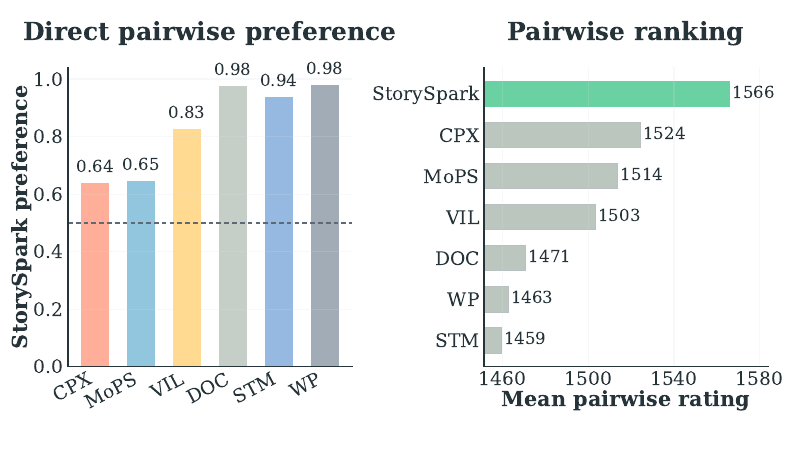}
  \vspace{-2em}
  \caption{
  Pairwise preference validation. Left: direct StorySpark preference score, where 0.5 indicates parity; right: mean candidate-level Elo rating.
  }
  \label{fig:pairwise_validation}
  \vspace{-1em}
\end{figure}

Figure~\ref{fig:pairwise_validation} evaluates the same premise candidates through relative preference rather than absolute rubric scores. StorySpark obtains a direct preference score above 0.5 against every baseline, meaning that it is preferred more often than it is dispreferred after ties are counted. The margins are largest against the external-source assets and VIL, while the scores against CPX and MoPS are more moderate. This pattern is useful: it shows that the strongest generation baselines remain competitive in some individual pairwise cases, but StorySpark still has a consistent advantage under direct comparison.

The candidate-level Elo ranking gives the same conclusion from the full tournament. StorySpark achieves the highest mean rating, ahead of CPX and MoPS, which are also the closest competitors in the direct preference view. Because the tournament uses adaptive rating-near pairing after warm-up rounds, many later comparisons focus on harder cross-method decisions rather than only obvious quality gaps. Thus, the pairwise results complement Table~\ref{tab:premise_quality}: StorySpark's gains are not only reflected in pointwise averages, but also persist when concrete premises are judged side by side.

\subsection{Robustness Across Judges and Evaluation Views}
\label{sec:robustness_validation}

\begin{table}[htbp]
\centering
\small
\setlength{\tabcolsep}{3.5pt}
\begin{tabular}{@{}lcc@{}}
\toprule
View & DeepSeek-V4-Flash & GPT-5.2 \\
\midrule
\multicolumn{3}{@{}l}{\textit{Main premise quality}} \\
Overall rank & 1st & 1st \\
Overall margin & +1.21 & +1.24 \\
Originality margin & +4.41 & +6.61 \\
\addlinespace[2pt]
\multicolumn{3}{@{}l}{\textit{Additional validation views}} \\
Story-transfer rank & 1st & 1st \\
Pairwise Elo rank & 1st & 2nd \\
\bottomrule
\end{tabular}
\caption{
Robustness across judges and views. Margins are against the strongest baseline; GPT-5.2 only rerates existing artifacts.
}
\vspace{-1em}
\label{tab:cross_judge_summary}
\end{table}

Table~\ref{tab:cross_judge_summary} summarizes robustness across judges and evaluation views. GPT-5.2 rerates the same main-premise artifacts without participating in generation, search, or story expansion, and StorySpark still ranks first: its Overall margin changes only from +1.21 to +1.24, while its Originality margin increases from +4.41 to +6.61. The auxiliary views show the same pattern: StorySpark ranks first in premise-to-story transfer under both judges, first in DeepSeek-V4-Flash pairwise Elo and second under GPT-5.2. These results indicate that the gains are not just a search-judge calibration artifact; StorySpark remains consistently strong, with the clearest signal in producing more original premise seeds.

\begingroup
\scriptsize
\setlength{\tabcolsep}{2pt}
\begin{table}[t]
\centering
\footnotesize
\setlength{\tabcolsep}{3.5pt}
\begin{tabular}{@{}lcccc@{}}
\toprule
Method & Fascination & Completeness & Originality & Overall \\
\midrule
StorySpark & 2.70 & \textbf{4.08} & \textbf{3.12} & \textbf{3.30} \\
MoPS       & \textbf{2.90} & \underline{4.04} & \underline{2.66} & \underline{3.20} \\
STM        & 2.15 & 2.54 & 1.88 & 2.19 \\
WP         & \underline{2.81} & 2.04 & 2.29 & 2.38 \\
\bottomrule
\end{tabular}
\caption{Human evaluation on anonymized story premises. Scores are averaged over four annotators on a 1--5 Likert scale; Overall is the mean of Fascination, Completeness, and Originality.}
\vspace{-1.5em}
\label{tab:human_eval}
\end{table}
\endgroup

Human evaluation provides a reader-facing check. Following the anonymized protocol in Section~\ref{sec:experiment_setup}, evaluators rate premises from the methods in Table~\ref{tab:human_eval} on Fascination, Completeness, and Originality using a 1--5 Likert scale. StorySpark obtains the best Overall score and ranks first on Completeness and Originality, while MoPS is slightly higher on Fascination. This mirrors the automatic results: StorySpark's advantage is strongest in producing more complete and original premise seeds.

\section{Conclusion}

StorySpark moves story premise generation from static modular composition to module-wise evolutionary search. It keeps the interpretable MoPS-style module structure, evaluates partial premises across fascination, completeness, and originality, and uses Pareto-guided local selection plus reserve-wildcard frontier allocation to balance quality, complementarity, and branch coverage. Across premise-level quality, ablations, process analysis, pairwise validation, story-level transfer, and embedding-based quality-aware diversity, the strongest recurring signal is improved originality while retaining completeness. Looking ahead, we hope to explore lower-cost intermediate evaluation, larger-scale creator feedback, interactive long-form planning, and richer narrative constraints, while keeping premise generation as a controllable early-stage aid for human writers.

\clearpage

\section*{Limitations}

\textbf{Fixed modular premise structure.}
StorySpark follows the MoPS-style premise schema, constructing a premise through ordered modules such as background, persona, event, ending, and twist. This fixed structure makes module-wise search interpretable, but story ideation is not always strictly linear. A later event, ending, or twist may suggest that an earlier background or persona should be revised. Future work will explore more dynamic module structures, including adaptive module ordering and backward updates from downstream modules to upstream ones.

\textbf{Larger-scale evaluation and human-guided evolution.}
StorySpark is evaluated through automatic judging, pairwise validation, story expansion, diversity analysis, and human evaluation. These views provide complementary evidence, but creative premise quality still depends on reader preference, author intent, and literary judgment. Future work will conduct larger-scale evaluations with readers, writers, and experts. We also plan to incorporate sparse human feedback into the evolutionary loop, so that a small amount of high-value human judgment can better guide premise search.


\section*{Ethical considerations}
StorySpark is designed as an early-stage creative ideation aid, not an autonomous replacement for human writers. Since it relies on LLM generation and automatic evaluation, its outputs may inherit biases, stereotypes, unsafe associations, or sensational narrative patterns from the underlying models and prompts, which may be further amplified by evolutionary search if favored by automatic judges. Generated premises and stories should therefore be treated as draft suggestions and reviewed by humans before publication or deployment. Although our evaluation considers premise quality, story expansion, pairwise preference, diversity, and human ratings, these signals are not definitive measures of literary value, which depends on reader background, cultural context, authorial intent, and genre conventions. Practical systems based on StorySpark should include human oversight, content moderation, and safeguards, especially for stories involving sensitive social groups, traumatic events, violence, or other potentially harmful themes.

\bibliography{refs}

\clearpage
\appendix

\providecommand{\AppTableCaption}[2]{%
  \refstepcounter{table}\label{#1}%
  \vspace{0.25em}\noindent{\footnotesize\textbf{Table~\thetable:} #2\par}\vspace{0.55em}%
}
\providecommand{\AppFigureCaption}[2]{%
  \refstepcounter{figure}\label{#1}%
  \vspace{0.25em}\noindent{\footnotesize\textbf{Figure~\thefigure:} #2\par}\vspace{0.55em}%
}
\providecommand{\AppAlgorithmCaption}[2]{%
  \refstepcounter{algorithm}\label{#1}%
  \noindent{\footnotesize\textbf{Algorithm~\thealgorithm} #2\par}%
}

\section{Algorithm Details}
\label{app:algorithm_details}
\label{app:algorithm}

This appendix reports the implementation details of StorySpark that are only summarized in the main text. The theme is treated as a fixed root condition, and the five active modules are background, persona, event, ending, and twist. We use \emph{branch quota} and \emph{shared quota} as implementation terms corresponding to the reserve and wildcard allocation described in Section~3.3.

\subsection{Default Module Budgets}
\label{app:default_budgets}

\begin{center}
\scriptsize
\begin{tabular}{@{}lrrrr@{}}
\toprule
Module & $B_t$ & $R_t$ & $G_t$ & $\mu_t$ \\
\midrule
Background & 6 & 6 & 18 & 0.7 \\
Persona    & 4 & 2 & 12 & 0.7 \\
Event      & 3 & 2 & 9  & 0.7 \\
Ending     & 2 & 1 & 6  & 0.7 \\
Twist      & 2 & 1 & 6  & 0.7 \\
\bottomrule
\end{tabular}
\end{center}
\AppTableCaption{tab:module_hyperparameters}{Default module-level hyperparameters. $B_t$ is the per-branch frontier budget, $R_t$ is the branch quota, $G_t$ is the target local-pool size after evolution, and $\mu_t$ is the probability of applying mutation rather than crossover.}

\subsection{Full Procedure}
\label{app:full_algorithm}

\smallskip
\noindent\begin{minipage}{\columnwidth}
\AppAlgorithmCaption{alg:storyspark_full}{StorySpark module-wise evolutionary search.}
\vspace{0.2em}
\hrule
\vspace{0.25em}
\scriptsize
\begin{algorithmic}[1]
\Require Theme $\tau$, modules $m_{1:5}$, budgets $\{B_t,R_t,G_t,\mu_t\}_{t=1}^{5}$
\Ensure Final scored premises $\mathcal{Y}^{\tau}$
\State $\mathcal{F}_0^{\tau}\gets\{\emptyset\}$
\For{$t=1$ to $5$}
  \State $\mathcal{F}_t^{\tau}\gets\emptyset$; $\mathcal{U}_t\gets\emptyset$
  \ForAll{$x\in\mathcal{F}_{t-1}^{\tau}$}
    \State $\mathcal{P}\gets\Call{InitAndScore}{\tau,x,m_t,B_t}$
    \State compute Pareto ranks and local crowding in $\mathcal{P}$
    \While{$|\mathcal{P}|<G_t$}
      \State sample parent candidate(s) from $\mathcal{P}$ using $w(c)$
      \State $c'\gets\Call{MutateOrCross}{\tau,x,m_t,\mathcal{P},\mu_t}$
      \If{$c'$ passes validity and duplicate checks}
        \State $z'\gets\Call{Synthesize}{\tau,x,m_t,c'}$
        \If{$z'$ is successfully materialized}
          \State $(\mathbf{q}',\rho')\gets\Call{Judge}{z'}$
          \State add $(c',z',\mathbf{q}',\rho')$ to $\mathcal{P}$
          \State update Pareto ranks and local crowding
        \EndIf
      \EndIf
    \EndWhile
    \State $\mathcal{Q}\gets\Call{SelectBranchQuota}{\mathcal{P},R_t}$
    \State add $\Call{Attach}{x,\mathcal{Q}}$ to $\mathcal{F}_t^{\tau}$
    \State add $\Call{RunnerUps}{\mathcal{P},\mathcal{Q},x}$ to $\mathcal{U}_t$
  \EndFor
  \State $H_t\gets(B_t-R_t)|\mathcal{F}_{t-1}^{\tau}|$
  \State $\mathcal{S}\gets\Call{SelectSharedQuota}{\mathcal{U}_t,H_t}$
  \State add $\Call{AttachStoredPrefixes}{\mathcal{S}}$ to $\mathcal{F}_t^{\tau}$
\EndFor
\State $\mathcal{Y}^{\tau}\gets\emptyset$
\ForAll{$x_{1:5}\in\mathcal{F}_5^{\tau}$}
  \State $y\gets\Call{Synthesize}{\tau,x_{1:5}}$
  \State $(\mathbf{q}(y),\rho(y))\gets\Call{Judge}{y}$
  \State add $(y,\mathbf{q}(y),\rho(y))$ to $\mathcal{Y}^{\tau}$
\EndFor
\State \Return $\mathcal{Y}^{\tau}$
\end{algorithmic}
\vspace{0.25em}
\hrule
\end{minipage}
\smallskip

Algorithm~\ref{alg:storyspark_full} shows the full module-wise search loop. A local-pool item stores the current-module candidate, its materialized partial premise, its quality vector, and the corresponding judge feedback. A shared-pool item additionally stores the parent-branch identity, so that shared-quota selections are attached back to the correct prefix rather than recombined across branches.

\subsection{Validity Filtering}
\label{app:validity_filtering}

Before a generated candidate enters a local pool, StorySpark applies four checks. First, the normalized candidate must contain non-empty current-module text. Second, it must not duplicate an existing candidate in the same local pool under normalized string matching. Third, for typed modules, it must satisfy the predefined module-type constraint. Fourth, it must be successfully materialized into a partial premise before scoring. Candidates that fail any check are discarded before Pareto ranking or frontier allocation.

\subsection{Frontier Allocation Implementation Notes}
\label{app:quota_selection}

The main text defines the branch-quota and shared-quota sorting keys. In implementation, branch-quota selection is performed independently inside each local pool, whereas shared-quota selection is performed over the runner-up candidates collected from all parent branches at the same module depth. Shared-pool candidates keep both their original parent-prefix identifier and their local-crowding value. This allows the shared quota to introduce cross-branch competition while still attaching each selected candidate back to its own prefix.

Pareto ranks used for branch-quota selection are computed within each local pool. Global Pareto ranks used for shared-quota selection are recomputed only over the shared runner-up pool at the current module. Deterministic tie-breaking is applied only as a reproducibility device under a fixed generation trace; it does not act as an additional quality signal.

\section{Prompt Templates}
\label{app:prompts}

This section keeps only the five StorySpark templates used by our method: initialization, crossover, mutation, module-to-premise synthesis, and scoring. Baseline prompts, pairwise-evaluation prompts, and story-expansion prompts are omitted to keep the appendix focused on the proposed search method.
\subsection{Initialization Prompt}
\label{app:init_prompt}

\begin{promptbox}{Initialization Prompt}
You are helping construct a modular story premise. The current task is to
generate candidates for the current slot. The fixed theme and any upstream slots
must not be rewritten.

\noindent\textbf{Theme:} \{\texttt{theme}\}\\
\textbf{Available upstream slots:} \{\texttt{prefix\_slots}\}\\
\textbf{Current slot:} \{\texttt{slot\_name}\}\\
\textbf{Number of candidates:} \{\texttt{target\_count}\}

Generate exactly \{\texttt{target\_count}\} candidates for the current slot.
Each candidate should fill only the current slot and remain compatible with the
theme and upstream slots.

For typed slots, each candidate must choose exactly one valid type:
\{\texttt{slot\_type\_options}\}

Return JSON only. For typed slots:
\begin{verbatim}
{
  "candidates": [
    {"text": "candidate 1", "type": "one_valid_type"},
    {"text": "candidate 2", "type": "one_valid_type"}
  ]
}
\end{verbatim}

For untyped slots:
\begin{verbatim}
{
  "candidates": [
    {"text": "candidate 1"},
    {"text": "candidate 2"}
  ]
}
\end{verbatim}
\end{promptbox}

\subsection{Crossover Prompt}
\label{app:crossover_prompt}

\begin{promptbox}{Crossover Prompt}
You are creating one new candidate in a modular story premise by combining two
parent candidates. The fixed context below must not be changed. Create only a
new \{\texttt{slot\_name}\} candidate.

\noindent\textbf{Current slot:} \{\texttt{slot\_name}\}\\
\textbf{Fixed context:} \{\texttt{context}\}

\noindent\textbf{Parent A:} \{\texttt{parent\_a}\}\\
\textbf{Feedback for Parent A:}\\
Fascination: \{\texttt{fascination\_reasoning\_a}\}\\
Completeness: \{\texttt{completeness\_reasoning\_a}\}\\
Originality: \{\texttt{originality\_reasoning\_a}\}

\noindent\textbf{Parent B:} \{\texttt{parent\_b}\}\\
\textbf{Feedback for Parent B:}\\
Fascination: \{\texttt{fascination\_reasoning\_b}\}\\
Completeness: \{\texttt{completeness\_reasoning\_b}\}\\
Originality: \{\texttt{originality\_reasoning\_b}\}

\noindent\textbf{Existing candidates that should not be repeated:}\\
\{\texttt{forbidden\_block}\}

Requirements:
\begin{itemize}
    \item Combine strengths from both parents into one new current-slot candidate.
    \item Keep the child concise and coherent with the fixed context.
    \item Do not repeat or trivially rephrase any existing candidate.
    \item For typed slots, return one valid candidate type.
\end{itemize}

Return JSON only:
\{"text": "new candidate", "type": "one\_valid\_type"\}

For untyped slots, omit \texttt{type}.
\end{promptbox}

\subsection{Mutation Prompt}
\label{app:mutation_prompt}

\begin{promptbox}{Mutation Prompt}
You are improving one candidate in a modular story premise. The fixed context
below must not be changed. Revise only the current
\{\texttt{slot\_name}\} candidate.

\noindent\textbf{Current slot:} \{\texttt{slot\_name}\}\\
\textbf{Fixed context:} \{\texttt{context}\}\\
\textbf{Parent candidate:} \{\texttt{parent\_text}\}

\noindent\textbf{Evaluation feedback for the parent candidate:}\\
Fascination: \{\texttt{fascination\_reasoning}\}\\
Completeness: \{\texttt{completeness\_reasoning}\}\\
Originality: \{\texttt{originality\_reasoning}\}

\noindent\textbf{Existing candidates that should not be repeated:}\\
\{\texttt{forbidden\_block}\}

Requirements:
\begin{itemize}
    \item Produce exactly one new candidate for the current slot.
    \item Make the new candidate meaningfully different from the parent.
    \item Keep it coherent with the fixed context.
    \item Do not repeat or trivially rephrase any existing candidate.
    \item For typed slots, consider the valid candidate types and return one type.
\end{itemize}

Return JSON only:
\{"text": "new candidate", "type": "one\_valid\_type"\}

For untyped slots, omit \texttt{type}.
\end{promptbox}

\subsection{Module-to-Premise Synthesis Prompt}
\label{app:module_to_premise_prompt}

\begin{promptbox}{Module-to-Premise Synthesis Prompt}
TASK: \{\texttt{PARTIAL\_PREMISE\_SYNTHESIS or FINAL\_SYNTHESIS}\}

The following story elements are currently available:

\noindent\textbf{Theme:} \{\texttt{theme}\}\\
\textbf{Background:} \{\texttt{background}\}\\
\textbf{Persona:} \{\texttt{persona}\}\\
\textbf{Event:} \{\texttt{event}\}\\
\textbf{Ending:} \{\texttt{ending}\}\\
\textbf{Twist:} \{\texttt{twist}\}

Please combine the available elements into one compact, concise, and coherent
sentence as a story premise. For partial synthesis, do not add specific details
for downstream slots that have not been provided yet. For final synthesis, use
the full theme, background, persona, event, ending, and twist.
\end{promptbox}

\subsection{Scoring Prompt}
\label{app:scoring_prompt}

\begin{promptbox}{Scoring Prompt}
Here is a story premise:

\{\texttt{premise}\}

Evaluate the story premise on the following three dimensions. Use a score from
0 to 100 for each dimension.

\noindent\textbf{Fascination:} Judge how engaging the premise is and how
strongly it makes you want to read the story developed from it.\\
\textbf{Completeness:} Judge whether the premise contains the essential elements
expected in a complete story premise.\\
\textbf{Originality:} Judge how familiar or novel the premise feels. For
originality, score the candidate only if it is at least a complete story
premise. Otherwise, give originality a score of 0.

Requirement: provide deterministic scores and concise explanations. For each
dimension, write the explanation before the numeric score.

Return JSON only:
\{
  "fascination": \{
    "explanation": "short explanation",
    "score": 84
  \},
  "completeness": \{
    "explanation": "short explanation",
    "score": 92
  \},
  "originality": \{
    "explanation": "short explanation",
    "score": 78
  \}
\}
\end{promptbox}

\section{Experimental Setup Details}
\label{app:experimental_setup}
\label{app:setup}
\label{app:extended_setup}
\label{app:implementation_details}

This appendix provides operational details for the experiment setup in Section ~\ref{sec:experiment_setup}, including model backends, comparison assets, output counts, evaluation units, artifact reuse, pairwise aggregation, story transfer, human evaluation, and quality-conditioned coverage.

\subsection{Model Backends and Decoding}
\label{app:model_decoding}

\begin{center}
\scriptsize
\resizebox{\columnwidth}{!}{%
\begin{tabular}{@{}lll@{}}
\toprule
Component & Backend / decoding & Used for \\
\midrule
Initialization & DeepSeek-V4-Flash, temp. 0.7 & Current-module candidate generation \\
Mutation & DeepSeek-V4-Flash, temp. 0.7 & Feedback-guided current-module revision \\
Crossover & DeepSeek-V4-Flash, temp. 0.7 & Current-module candidate recombination \\
Partial-premise synthesis & DeepSeek-V4-Flash, temp. 0.7 & Contextual materialization for scoring \\
Final-premise synthesis & DeepSeek-V4-Flash, temp. 0.7 & Complete premise generation \\
Primary pointwise judge & DeepSeek-V4-Flash, det. & Search feedback and main automatic scoring \\
Independent judge & GPT-5.2, det. & Main-artifact rerating \\
Pairwise judge & DeepSeek-V4-Flash, det. & A/B/TIE pairwise validation \\
Story writer & DeepSeek-V4-Flash, temp. 0.7 & Premise-to-story transfer \\
Embedding model & Qwen/Qwen3-Embedding-4B & Local crowding and semantic coverage \\
\bottomrule
\end{tabular}}
\end{center}
\AppTableCaption{tab:model_decoding}{Model and decoding settings. ``det.'' denotes deterministic decoding. Generation-style calls use temperature 0.7, while evaluation-style calls are deterministic.}

Structured outputs are parsed after each call. If a response cannot be parsed into the expected fields, the same prompt is retried within the implementation retry budget; if parsing still fails, the output is discarded. Candidate-level failures do not terminate a run. Duplicate candidates are removed by normalized string matching within each local pool.

\subsection{Genre Domains, Assets, and Output Counts}
\label{app:genre_assets}

Experiments use the 14 genre domains from MoPS: Fantasy, Fantastic, Martial Arts, Immortal Heroes, Urban, Contemporary, Military, Historical, Game, Sports, Science Fiction, Suspense, Romance, and Time-travel. Baseline definitions follow Section~4.1; this appendix only records the asset handling and output-count details needed to reproduce the experiments.

\noindent\textbf{StorySpark output count.}
StorySpark searches over five active modules: background, persona, event, ending, and twist. The promoted branching factor at each module is the frontier budget $B_t$, so the number of complete module paths per genre domain is determined by the product of these promoted factors:

\begin{center}
\scriptsize
\resizebox{\columnwidth}{!}{%
\begin{tabular}{@{}lccccc@{}}
\toprule
Module & Background & Persona & Event & Ending & Twist \\
\midrule
Promoted factor $B_t$ & 6 & 4 & 3 & 2 & 2 \\
Cumulative paths & 6 & 24 & 72 & 144 & 288 \\
\bottomrule
\end{tabular}}
\end{center}
\AppTableCaption{tab:output_count}{Promoted branching factors and cumulative final-path counts for one genre domain.}

Thus, one genre domain yields
\[
6 \times 4 \times 3 \times 2 \times 2 = 288
\]
complete module paths, and one full StorySpark run over all 14 domains produces
\[
14 \times 288 = 4{,}032
\]
final premises. The branch quota $R_t$ and local-pool target $G_t$ affect within-module preservation, competition, and evolution, but they do not change the final number of promoted complete paths.

\noindent\textbf{Generated comparison outputs.}
MoPS, VIL, and CPX are generated comparison systems. VIL directly generates complete premises, while CPX follows the Complex baseline setting in MoPS by conditioning direct prompting on fixed example premises. These examples are fixed across runs and are not selected from StorySpark outputs.

\noindent\textbf{External-source assets.}
DOC, WritingPrompts, and Storium are taken from the released MoPS baseline assets. We do not recollect, regenerate, or re-filter the original external datasets; only format normalization required by our evaluation scripts is applied. These external-source assets are evaluated as aggregate premise sets rather than genre-by-genre generated systems.

\subsection{Evaluation Units and Sample Counts}
\label{app:sample_counts}

\begin{center}
\scriptsize
\resizebox{\columnwidth}{!}{%
\begin{tabular}{@{}llll@{}}
\toprule
Evaluation & Unit & Methods / artifacts & Sample count \\
\midrule
Main premise comparison & Final premise & All seven methods & 1,000 per method \\
Quality-conditioned coverage & Final-premise embeddings & Same main artifacts & 1,000 per method \\
Independent rerating & Final premise & Same main artifacts & 1,000 per method \\
Mechanism ablation & Final premise & StorySpark variants & 4,032 per variant \\
module-depth process analysis & Partial premise & Full search traces & All retained partial states \\
Pairwise validation & Premise pair & All seven methods & 100 candidates per method; 5,600 primary pairs + 5,600 audits \\
Premise-to-story transfer & Expanded story & All seven methods & 100 per method \\
Human evaluation & Final premise & StorySpark, MoPS, STM, WP & 20 per method \\
\bottomrule
\end{tabular}}
\end{center}
\AppTableCaption{tab:sample_counts}{Evaluation units and sample counts used in the main experiments.}

The main comparison uses equal-sized final-premise samples from all seven methods. The independent automatic judge rerates the same main-comparison artifacts rather than regenerated outputs. Mechanism ablations evaluate full output sets from each StorySpark variant. module-depth process analysis scores materialized partial premises retained during search, so it evaluates intermediate construction states rather than only final premises.



\subsection{Pointwise Scoring and Artifact Reuse}
\label{app:pointwise_reuse}

The pointwise dimensions and score scales are defined in Section~4.2. Here we clarify how these scores are used during search and how evaluation artifacts are reused across judges.

\noindent\textbf{Search-time feedback.}
During StorySpark search, the primary pointwise judge scores each materialized partial premise and returns concise feedback. The score vector is used for Pareto ranking, while the feedback is passed to mutation and crossover prompts. This feedback is only used inside the StorySpark search loop and is not available to baseline generation systems.

\noindent\textbf{Paper-facing evaluation artifacts.}
For final reporting, all generated artifacts are frozen before evaluation. The main premise comparison evaluates the same final-premise sample for each method; the story-transfer experiment first expands the sampled premises with a fixed writer and then scores the resulting stories. Overall is computed as the arithmetic mean of fascination, completeness, and originality.

\noindent\textbf{Independent rerating.}
GPT-5.2 is used only as an independent evaluator. It rerates existing final premises, expanded stories, and anonymized pairwise comparisons without regenerating premises, modifying prompts, rewriting stories, or changing baseline outputs. Therefore, differences between DeepSeek-V4-Flash and GPT-5.2 results reflect evaluator variation rather than new generation runs.

\subsection{Pairwise Validation and Elo Aggregation}
\label{app:pairwise_protocol}

Pairwise validation is constructed from the main premise-evaluation artifacts. We sample 100 premises per method from StorySpark, CPX, MoPS, VIL, DOC, Storium, and WritingPrompts, yielding 700 candidate premises in total. Sampling uses a fixed random seed of 42. When method outputs contain theme labels, we apply approximately theme-balanced sampling by allocating near-uniform quotas across themes; otherwise, we fall back to random sampling.

\noindent\textbf{Pair construction.}
Each candidate participates in 16 primary cross-method comparisons. The first two rounds use random warm-up pairing across different methods. The remaining 14 rounds use rating-near adaptive pairing: candidates are paired with candidates from other methods whose current Elo ratings are close. This makes later comparisons focus on more contested decisions rather than only obvious quality gaps. We exclude within-method comparisons and do not repeat the same candidate pair.

With 700 candidates and 16 primary comparisons per candidate, the primary tournament contains
\[
700 \times 16 / 2 = 5{,}600
\]
unordered candidate pairs. Each primary comparison is then audited once with the presentation order reversed, resulting in another 5,600 judge calls.

\noindent\textbf{Judge protocol and order audit.}
For each comparison, the judge sees two anonymized premises, labeled A and B, and returns one of A, B, or TIE. The prompt asks the judge to assess overall premise quality in terms of fascination, completeness, expandability, originality, and internal consistency. It also instructs the judge not to decide based on length, surface writing style, formatting, genre preference, or whether the premise appears human- or machine-written. The output is parsed into a winner field and a short reason.

To reduce positional bias, every comparison is repeated with the candidate order reversed. After mapping the reversed judgment back to the original candidate identities, we keep the result if the forward and reversed judgments agree. If they conflict, the final outcome for that candidate pair is converted to TIE. In the reported run, 3,269 primary comparisons remain direction-consistent and 2,331 are converted to ties by the reverse-order audit.

\noindent\textbf{Direct StorySpark preference.}
The direct preference score in Figure~5 is computed only from direct StorySpark-vs.-baseline encounters. For a comparison between StorySpark and baseline $b$, StorySpark receives score
\[
s =
\begin{cases}
1, & \text{StorySpark wins},\\
0.5, & \text{TIE},\\
0, & \text{StorySpark loses}.
\end{cases}
\]
The direct preference score against baseline $b$ is
\[
\mathrm{Pref}(\mathrm{StorySpark}, b)
=
\frac{1}{N_b}\sum_{j=1}^{N_b} s_j .
\]
This score is not a raw win rate: a value of 0.5 indicates parity, while a value above 0.5 indicates that StorySpark is preferred more often in direct comparisons. Because later rounds use adaptive rating-near pairing, $N_b$ differs across baselines; StorySpark is compared more often with closer competitors such as CPX and MoPS.

\begin{center}
\scriptsize
\begin{tabular}{@{}lrrrrrr@{}}
\toprule
Baseline $b$ & CPX & MoPS & VIL & DOC & STM & WP \\
\midrule
Direct pairs $N_b$ & 767 & 460 & 238 & 42 & 47 & 46 \\
\bottomrule
\end{tabular}
\end{center}
\AppTableCaption{tab:direct_pair_counts}{Number of direct StorySpark-vs.-baseline comparisons in the reported adaptive pairwise run.}

\noindent\textbf{Candidate-level Elo ranking.}
For the global pairwise ranking, each premise candidate is treated as one Elo player. All candidates start with rating $R_0=1500$, and we use update factor $K=24$. For a comparison between candidates A and B, we assign
\[
s_A =
\begin{cases}
1, & \text{A wins},\\
0.5, & \text{TIE},\\
0, & \text{B wins}.
\end{cases}
\]
The expected score for A is
\[
E_A =
\frac{1}{1 + 10^{(R_B - R_A)/400}}.
\]
Ratings are updated by
\[
\begin{aligned}
R_A &\leftarrow R_A + K(s_A - E_A), \\
R_B &\leftarrow R_B + K\big((1-s_A) - (1-E_A)\big).
\end{aligned}
\]
The method-level pairwise rating is the mean final Elo rating over the 100 sampled candidates from that method:
\[
\overline{R}_m =
\frac{1}{|C_m|}
\sum_{c \in C_m} R_c,
\]
where $C_m$ is the set of sampled candidates from method $m$. Thus, the left panel of Figure~5 measures direct StorySpark-vs.-baseline preference, while the right panel summarizes all adaptive pairwise comparisons through candidate-level Elo aggregation.

\subsection{Premise-to-Story Transfer}
\label{app:story_transfer_protocol}

For premise-to-story transfer, we sample 100 premises per method and expand each premise with the same fixed StoryWriter-style pipeline. The story writer backend and decoding settings are held constant across methods. The resulting stories are then scored with the same three pointwise quality dimensions as premises. This protocol tests whether differences among premise seeds carry over to downstream story generation.

\subsection{Human Evaluation Protocol}
\label{app:human_eval_protocol}

Human evaluation uses anonymized final premises from StorySpark, MoPS, Storium, and WritingPrompts. We sample 20 premises per method, yielding 80 premises in total. Four evaluators rate each premise on fascination, completeness, and originality using a 1--5 Likert scale. Method names are hidden and premise order is randomized independently for each evaluator. 

Each evaluator received a spreadsheet containing the randomized premises and rating fields. The spreadsheet presents one premise per row. Evaluators see only an anonymized item identifier, the premise text, and three Likert-scale rating columns; they do not see the source method. Table~\ref{tab:human_eval_sheet_example} shows the structure of the annotation sheet. The instruction block asks evaluators to judge each premise independently, where 1 indicates very poor quality, 3 indicates acceptable or mixed quality, and 5 indicates excellent quality for the corresponding criterion.

\begin{table}[t]
\centering
\small
\resizebox{0.86\linewidth}{!}{%
\begin{tabular}{@{}lp{0.52\linewidth}ccc@{}}
\toprule
\textbf{Item ID} & \textbf{Premise} & \textbf{Fasc.} & \textbf{Comp.} & \textbf{Orig.} \\
\midrule
E037 &
A cartographer finds blank map regions spreading into real towns. &
1--5 & 1--5 & 1--5 \\
E012 &
A child interprets a village's shared recurring nightmare. &
1--5 & 1--5 & 1--5 \\
E064 &
A ship worker discovers letters from unborn passengers. &
1--5 & 1--5 & 1--5 \\
\bottomrule
\end{tabular}%
}
\caption{Example structure of the human evaluation spreadsheet. The examples illustrate the fields visible to evaluators; source method names are omitted and item order is randomized.}
\label{tab:human_eval_sheet_example}
\end{table}

\paragraph{Evaluator recruitment and payment.}
The four evaluators were adult students recruited from the authors' peer network. They were not authors of this paper and were not involved in developing StorySpark, generating the evaluated premises, or selecting the evaluated examples. The evaluators consisted of one female evaluator and three male evaluators. We did not collect other demographic attributes, such as country of residence, age, ethnicity, or socioeconomic background.

Participation was voluntary, and no monetary compensation was provided. The evaluation was conducted as a small-scale internal research assessment rather than a paid crowdsourcing task. Each evaluator rated 80 short premises, with three Likert-scale ratings per premise. The workload was therefore limited in scope and did not require specialized equipment, long-term commitment, or repeated sessions. No personally identifying information is reported; only aggregate rating results and aggregate gender distribution are included in the paper. Evaluators were informed that their ratings would be used only in aggregate form for research reporting.

\subsection{Quality-Conditioned Coverage}
\label{app:coverage_detail}

The main diversity analysis follows the MoPS-style breadth view but conditions it on automatic quality. For each method $m$ and Overall threshold $\tau$, let $P_m$ be the set of final premises and let
\[
P_m^{\ge \tau} = \{p \in P_m : \mathrm{Overall}(p) \ge \tau\}.
\]
We embed the retained premises with Qwen/\allowbreak Qwen3-\allowbreak Embedding-4B, project them into a two-dimensional t-SNE space, and compute the convex-hull area covered by their projected points. The reported quality-conditioned coverage is
\[
\mathrm{Coverage}_m(\tau)
=
\mathrm{Breadth}(P_m^{\ge \tau})
\cdot
\frac{|P_m^{\ge \tau}|}{|P_m|}.
\]
When no premises are retained at a threshold, coverage is set to zero. The main text reports thresholds $\tau \in \{80,82,84,86\}$. This weighting prevents a method from receiving high diversity credit only by retaining a small number of scattered high-scoring premises, and it also prevents low-quality outliers from inflating semantic breadth.

\FloatBarrier
\section{More Results}
\label{app:more_results}

\subsection{GPT-5.2 Cross-Judge Replication}
\label{app:gpt52_rerating}

\begin{figure}[H]
  \centering
  \includegraphics[width=\columnwidth]{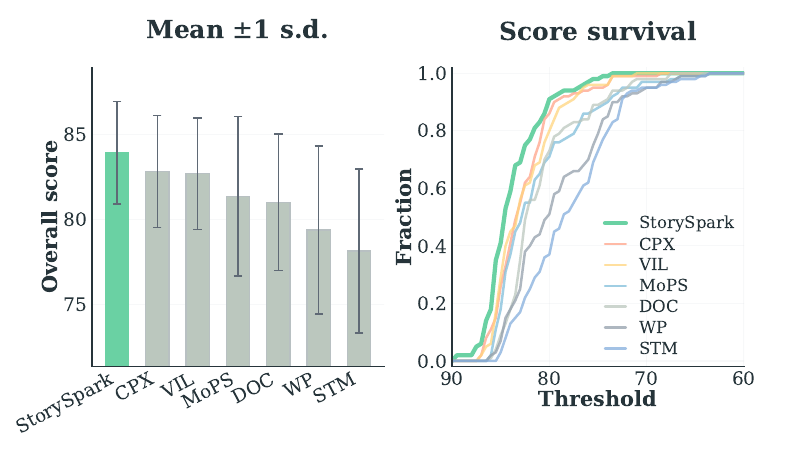}
  \vspace{-2em}
  \caption{GPT-5.2 cross-judge story-level transfer using the same expanded stories as the primary story-transfer experiment. Left: mean story Overall with \(\pm 1\) s.d.; right: score-survival curve over GPT-5.2 Overall thresholds.}
  \vspace{-1em}
  \label{fig:gpt52_story_transfer}
\end{figure}

Figure~\ref{fig:gpt52_story_transfer} shows that StorySpark remains the top method under GPT-5.2 story-level rerating. It achieves the highest mean Overall score and keeps the strongest high-score survival profile.

\begin{figure}[H]
  \centering
  \includegraphics[width=\columnwidth]{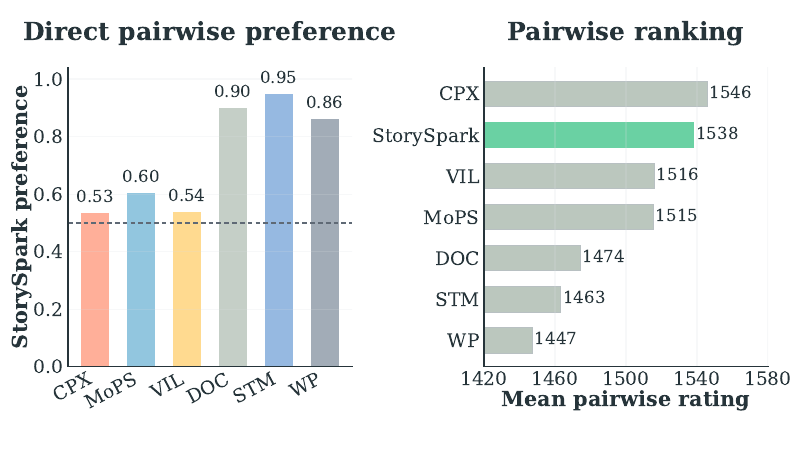}
  \vspace{-2em}
  \caption{GPT-5.2 cross-judge pairwise validation using the same anonymized premise pairs as the primary pairwise experiment. Left: direct StorySpark preference against each baseline; right: global pairwise ranking by mean candidate-level Elo rating.}
  \vspace{-1em}
  \label{fig:gpt52_pairwise}
\end{figure}

Figure~\ref{fig:gpt52_pairwise} shows that StorySpark remains favored in all direct StorySpark-vs-baseline comparisons under GPT-5.2. Its global Elo rank is second behind CPX, indicating that the direct preference trend is robust.

\FloatBarrier
\section{Case Study}
\label{app:case-study}

We present a same-theme case study under the \textit{Martial Arts} theme.
The comparison includes MoPS and StorySpark. For each method, we show the
module path, the synthesized premise, and the story-level scores after expansion
with the same fixed story writer.

\Needspace{14\baselineskip}
\subsection{Module Paths and Premises}
\label{app:case-module-premise}

\par\smallskip
\noindent\begin{minipage}{\columnwidth}
\centering
\scriptsize
\setlength{\tabcolsep}{3pt}
\renewcommand{\arraystretch}{1.08}
\begin{tabular}{@{}>{\raggedright\arraybackslash}p{0.22\linewidth}
                >{\raggedright\arraybackslash}p{0.72\linewidth}@{}}
\toprule
Field & Content \\
\midrule
Method
& MoPS \\

Theme
& Martial Arts \\

Background
& Qin imperial court, as seen in \textit{Hero}. \\

Persona
& A young idealistic court guard loyal to Qin martial discipline. \\

Event
& The guard discovers a hidden manual teaching a humane, fluid fighting style. \\

Ending
& The guard burns the manual and then trains from its ghost. \\

Twist
& The ghost is the guard's own future self, ensuring the book is burned and
history remains chained. \\

Premise
& A young, idealistic Qin court guard, whose rigid martial arts training embodies
absolute loyalty, discovers a forbidden manual teaching a fluid, humane style
suppressed by the court, and after burning the book, secretly trains from its
ghost---only to learn the ghost is their own future self, sent back to ensure
they burn it and live, so history's chains remain unbroken. \\
\bottomrule
\end{tabular}
{\captionsetup{hypcap=false}
\captionof{table}{MoPS module path and synthesized premise under the
\textit{Martial Arts} theme.}
\label{tab:case-mops-module-premise}}
\end{minipage}
\par\medskip

\par\smallskip
\noindent\begin{minipage}{\columnwidth}
\centering
\scriptsize
\setlength{\tabcolsep}{3pt}
\renewcommand{\arraystretch}{1.08}
\begin{tabular}{@{}>{\raggedright\arraybackslash}p{0.22\linewidth}
                >{\raggedright\arraybackslash}p{0.72\linewidth}@{}}
\toprule
Field & Content \\
\midrule
Method
& StorySpark \\

Theme
& Martial Arts \\

Background
& Ancient China during the Warring States period. \\

Persona
& A former court musician with broken hands who uses a modified guqin with
razor strings and qi resonance against a deaf warlord using silence and
null-qi fields. \\

Event
& The lost melody is not a song but the silences between notes. \\

Ending
& The protagonist and the warlord create harmonic silence that heals the land. \\

Twist
& The silences map the warlord's guilt: he was her student and broke her hands
to force the cure for his own affliction. \\

Premise
& A former court musician whose hands were broken by her deaf warlord student
wields a guqin that fires razor-edged strings and qi-disrupting vibrations to
confront his null-qi fields of silence, only to discover the lost melody---a
sequence of silences mapping his guilt---and, in a final stillness, plucks a
single string that vibrates through his bones, allowing him to counterpoint her
resonance with his own null-qi core, forging a harmonic silence that mends the
land and binds them as co-creators of a soundless peace born from their mutual
wounds. \\
\bottomrule
\end{tabular}
{\captionsetup{hypcap=false}
\captionof{table}{StorySpark module path and synthesized premise under the
\textit{Martial Arts} theme.}
\label{tab:case-storyspark-module-premise}}
\end{minipage}
\par\medskip

\Needspace{12\baselineskip}
\subsection{Story-level Scores}
\label{app:case-story-scores}

\par\smallskip
\noindent\begin{minipage}{\columnwidth}
\centering
\footnotesize
\setlength{\tabcolsep}{4pt}
\renewcommand{\arraystretch}{1.08}
\begin{tabular}{@{}lccc@{}}
\toprule
Method & Judge & Overall & F/C/O \\
\midrule
MoPS
& DeepSeek
& 83.33
& 85 / 90 / 75 \\

MoPS
& GPT-5.2
& 86.00
& 87 / 90 / 81 \\

StorySpark
& DeepSeek
& \textbf{88.33}
& \textbf{88} / \textbf{95} / \textbf{82} \\

StorySpark
& GPT-5.2
& \textbf{88.00}
& \textbf{88} / 90 / \textbf{86} \\
\bottomrule
\end{tabular}
{\captionsetup{hypcap=false}
\captionof{table}{Story-level scores for the two same-theme expanded stories.
F/C/O denotes fascination, completeness, and originality. MoPS story
contains six chapters and approximately 5,593 words; StorySpark story
contains six chapters and approximately 5,831 words.}
\label{tab:case-story-scores}}
\vspace{-1em}
\end{minipage}
\par\medskip

\Needspace{16\baselineskip}
\subsection{Expanded Stories}
\label{app:case-expanded-stories}

\storyblockstart
{Expanded MoPS story under the \textit{Martial Arts} theme.}

\begingroup
\footnotesize
\setlength{\parindent}{0pt}

\storyheading{Chapter 1: The Discovery}

\storypara{The dust of ages coated Wei Liang’s tongue. He had drawn the short straw, the unenviable duty of delivering a sealed scroll to the deepest recess of the Imperial Archives—a labyrinth of shelves and shadows forbidden to any guard of his rank. The air was thick with the scent of decaying parchment and cedar oil, a tomb for forgotten knowledge. He moved with the precise, economical steps drilled into him since childhood, his sword a silent weight at his hip, a symbol of his unwavering loyalty to the Qin code. Every step was a testament to the rigid, geometric forms he had mastered, forms that promised order and absolute obedience.}

\storypara{Navigating the dimly lit chamber, his boot caught on a flagstone that had shifted with age. He stumbled, his hand shooting out to brace against the cold stone wall. A loose brick, disguised by shadows, groaned and slid inward, revealing a dark cavity. A prickle of fear, quickly suppressed, ran down his spine. This was a breach, a flaw in the perfect architecture of the court. He should report it. He should seal it immediately.}

\storypara{But his hand, as if with a will of its own, reached into the void. His fingers brushed against silk, smooth and cool. He withdrew a folded bundle, its fabric a deep, riverine blue that seemed to drink the faint torchlight. Unfurling it, he saw characters painted in a flowing, cursive hand: “The Flowing River Style.” The title alone was a heresy. His martial training, the very foundation of his being, was built on the unyielding principles of stone and iron—strikes as straight as a plumb line, defenses as solid as a city wall. This manual spoke of empathy, of yielding, of adapting like water to the shape of its container.}

\storypara{He should burn it. He should march directly to his captain and report the seditious text. The Qin court tolerated no deviation. But a deep, unfamiliar curiosity, a hunger he did not know he possessed, gnawed at him. He glanced at the sealed scroll in his other hand. Duty could wait a single breath.}

\storypara{He turned a single page. The characters seemed to writhe before his eyes. He read of a stance that was not a stance, a movement that was not a strike, but a gentle redirection of an opponent’s force. As his mind struggled to comprehend the alien philosophy, a strange ache bloomed in his muscles, a deep, forgotten potential stirring in his bones. It was as if his own body recognized the truth of the words before his mind could.}

\storypara{Then, the ink shimmered. The black strokes lifted from the silk like mist, coalescing into a faint, spectral hand. It was translucent, the color of moonlight on water. It stretched from the page, slow and deliberate, and rested, feather-light, on his sword hand. A jolt, not of pain, but of profound recognition, shot through him. The touch was cold, yet familiar. A whisper, dry as the dust around them, brushed against his ear, a voice that was both alien and intimately known.}

\storypara{“Do not burn it, Wei Liang. Not yet.”}

\storypara{The ghostly hand faded, leaving only the thrum of his own panicked heart and the impossible weight of the silk manual in his hands. The sealed scroll lay forgotten at his feet. The decision was no longer a simple choice between duty and curiosity. It was a conversation with a ghost. He had to hide it. He had to know who—or what—had touched him. He quickly stuffed the manual back into the cavity, his fingers trembling, and pushed the brick back into place. The archives were silent once more, but the air was charged with a secret that now lived in his chest, a secret that had just touched his hand.}

\storyheading{Chapter 2: The Burning and the Oath}

\storypara{The flames had not died. They coiled within the brazier like living serpents, casting long, wavering shadows across the cracked flagstones of the courtyard. Wei Liang’s hands still trembled, though he had clenched them into fists until the knuckles shone white. The silk was gone. The forbidden words, the diagrams of impossible grace, the heresy of the Flowing River Style—all ash. He had done it. He had proven his loyalty.}

\storypara{But the air was wrong.}

\storypara{The smoke from the brazier did not rise and scatter as it should have. Instead, it pooled, thickened, and began to take shape. A chill crept up Wei Liang’s spine, a cold that had nothing to do with the winter night. He reached for his sword, but his hand froze mid-grip. The smoke coalesced into the outline of a man—taller than him, broader at the shoulder, with a face that was familiar in a way that made his stomach drop.}

\storypara{It was his own face.}

\storypara{The ghost wore the same simple guard’s tunic, but it was torn and stained with old, dark blood. Its eyes held a weariness that Wei Liang had never seen in a mirror, and its hair was streaked with grey at the temples. It stood within the haze of the dying fire, translucent as morning frost, and yet more real than the stars above.}

\storypara{“You did right,” the ghost said. Its voice was his, but deeper, cracked from disuse or grief. “Now, you must learn what you burned.”}

\storypara{Wei Liang’s breath caught. He took a step back, his heel scraping against stone. “You are the specter from the manual. I destroyed it. You should be gone.”}

\storypara{The ghost smiled, and it was a sad, knowing thing. “Fire cannot destroy what is written in the marrow. The silk is gone, but the words remain—in me, and now in the memory of your eyes. You read three lines before you threw it into the flames. That is enough to begin.”}

\storypara{“Begin what?” Wei Liang’s voice cracked. “I swore an oath. I will not betray the court.”}

\storypara{“You swore an oath to burn the book,” the ghost said, stepping closer. Its feet did not touch the ground. “You kept it. But the oath said nothing about forgetting. And I am here to ensure you do not forget.”}

\storypara{Wei Liang’s hand finally found his sword hilt, but he did not draw. The ghost was unarmed, and yet it radiated a quiet authority that made the blade feel like a child’s toy. “Who are you? Why do you wear my face?”}

\storypara{The ghost tilted its head, and for a moment, something like pain flickered in its eyes. “I am the memory of the Flowing River Style. I am every hand that turned its pages, every breath that shaped its forms. And I am you, Wei Liang—the you that will be, if you survive what is coming.”}

\storypara{“That is impossible.”}

\storypara{“Is it?” The ghost raised its hand, and the smoke from the brazier curled around its fingers like a living ribbon. “The manual was not written by men of this dynasty. It was written by a man who saw the chain of years as a river, not a wall. He learned to send his will backward, to whisper to those who came before. I am his echo, and yours.”}

\storypara{Wei Liang shook his head, but the denial died in his throat. The ghost’s presence was undeniable—a weight in the air, a pressure against his skin. He felt it in his bones, a resonance that hummed with the same rhythm as his own heartbeat.}

\storypara{“You burned the book,” the ghost continued, “because you were afraid. Good. Fear is the first gate. But now you must learn what you burned, not with your eyes, but with your body. The manual cannot be destroyed by fire. It can only be destroyed by being memorized, lived, and then forgotten. That is the only way to break the chain.”}

\storypara{“Break the chain?” Wei Liang’s voice was barely a whisper.}

\storypara{The ghost’s expression hardened. “The Qin court has bound itself with iron laws, and those laws have bound the spirit of every warrior who serves them. The Flowing River Style is not a weapon. It is a key. If you learn it, truly learn it, you will not be able to obey the court’s commands without question. You will see the chains for what they are. And you will have the strength to break them—or the wisdom to bear them.”}

\storypara{Wei Liang’s hand fell from his sword. “And if I refuse?”}

\storypara{The ghost’s eyes glowed faintly, like embers. “Then I will remain here, trapped between the flame and the ash, and you will live out your days as a perfect guard—loyal, obedient, and hollow. But the chain will not break. And in thirty years, when the court orders you to put down a village for the crime of feeding its children, you will do it. And you will remember this night, and you will wish you had learned.”}

\storypara{The silence stretched. The brazier crackled, the last embers dying into grey.}

\storypara{Wei Liang looked at the ghost—at his own future face, lined with sorrow and purpose. He thought of the oath he had sworn, the loyalty he had pledged. He thought of the words he had read, only three lines, but they had moved through him like water through dry earth.}

\storypara{“How long?” he asked.}

\storypara{The ghost’s smile returned, softer now. “Until you are whole.”}

\storypara{It extended its hand. The smoke parted, and for a moment, the hand was solid—warm, human, waiting.}

\storypara{Wei Liang took a breath. Then he reached out and clasped it.}

\storyheading{Chapter 3: The Secret Training}

\storypara{The first lesson came in a dream, though Wei Liang would later swear it was more real than any waking hour. He stood in a void that was not empty—a landscape of shifting shadows and echoes, where his past combat forms flickered like candle flames in a draft. Here, the ghost materialized not from smoke but from the very fabric of his memory, a darker shape against the dark.}

\storypara{“Forget the root,” the ghost said. Its voice was a whisper of wind through dry reeds. “Forget the stance. Forget the salute. The Qin court teaches you to stand like a tree—rigid, rooted, waiting to be broken. Water does not wait. Water does not root.”}

\storypara{The ghost moved, and Wei Liang saw a flow so seamless it seemed to bend the air around it. There was no beginning, no end—only motion that redirected, yielded, and struck from the yielding. The ghost’s hand brushed Wei Liang’s chest, and he felt no force, only a gentle pressure that somehow sent him stumbling backward.}

\storypara{“Again,” the ghost said.}

\storypara{Night after night, they trained. Wei Liang learned to feel the intent behind a strike before the strike was thrown—a subtle tension in the shoulder, a shift of weight, a breath held too long. He learned to turn an opponent’s momentum against itself, to flow around a blow like water around a stone. His body began to move in ways that surprised him, muscles remembering patterns his mind had never been taught.}

\storypara{But the waking hours grew strained. In the barracks, his senior officers noticed. Captain Zhao, a man whose face was a mask of perpetual disapproval, pulled him aside after a sparring session.}

\storypara{“You’re quicker,” Zhao said, his eyes narrowing. “But you hesitate. I saw it. You had an opening to strike the throat, and you pulled back. Why?”}

\storypara{Wei Liang’s heart hammered. “I didn’t want to injure my training partner, sir.”}

\storypara{“Injury is the price of readiness,” Zhao said coldly. “You are a guard of the Qin court. Your duty is not to hesitate. It is to obey, to strike, to end.”}

\storypara{That night, Wei Liang entered the void with the ghost’s words and Zhao’s words warring in his mind. The ghost waited, patient as still water.}

\storypara{“Your captain is afraid,” the ghost said. “He sees you changing. He sees the chains loosening.”}

\storypara{“He sees a weakness,” Wei Liang replied.}

\storypara{“He sees a threat.” The ghost extended its hand. “Come. Show me what you have learned.”}

\storypara{They moved through the void, shadow against shadow, and Wei Liang felt something shift inside him. His body remembered the ghost’s teachings, and for the first time, he did not think. He simply flowed. The ghost’s attacks came faster, sharper, but Wei Liang slipped through them like a current through reeds. He felt the ghost’s intent before each move—the flicker of purpose, the breath of decision.}

\storypara{And then, he touched it.}

\storypara{His hand found the ghost’s shoulder, light as a falling leaf. It was the first time he had landed a strike in their training. The ghost froze, and for a split second, its face solidified.}

\storypara{It was his own face.}

\storypara{Aged. Scarred. Lines ran like dry riverbeds across the cheeks, the brow, the jaw. The eyes held a thousand silences—a thousand things seen, a thousand things unsaid, a thousand things that could not be undone.}

\storypara{Wei Liang snatched his hand back as though burned. “You… you are me.”}

\storypara{The ghost’s face dissolved back into shadow, but its voice was raw, worn thin by time. “I was you. I am what you will become if you learn—and what you will not become if you refuse.”}

\storypara{“Why do you wear my face?” Wei Liang’s voice cracked.}

\storypara{The ghost was silent for a long moment. Then, softly: “Because the river flows both ways. And I have seen where it ends.”}

\storypara{Wei Liang fell to his knees in the void. The shadows swirled around him, whispering echoes of his past forms—the stances of a loyal guard, the strikes of a man who had never questioned. He looked at his hands. They were the same hands that had held the burning scroll. The same hands that had touched a ghost.}

\storypara{“I don’t want to be you,” he said, his voice barely a whisper.}

\storypara{“Then learn to be more,” the ghost replied. “But know this: the scars you see are not from battle. They are from silence. From watching. From knowing what comes and being unable to stop it.”}

\storypara{The void trembled. The ghost’s form began to fray at the edges, dissolving into the shifting shadows. Wei Liang reached out, but his hand passed through empty air.}

\storypara{“Wait,” he said. “I still don’t understand.”}

\storypara{“You will,” the ghost’s voice came, fainter now. “When the river carries you to the same shore. But for now—train. And do not tell them what you have seen.”}

\storypara{The dream shattered. Wei Liang woke in his cot, gasping, his hand still outstretched. The dawn light crept through the barracks window, pale and cold. He looked at his palm, half-expecting to see the ghost’s touch still lingering there.}

\storypara{It was empty.}

\storypara{But his heart was not. The face—his face, scarred and ancient—burned behind his eyes. The ghost had not lied. It had shown him a truth he was not ready to accept. And somewhere deep in the currents of his mind, he felt the river begin to flow, carrying him toward a future he could not yet see, but could no longer deny.}

\storyheading{Chapter 4: The Broken Seal}

\storypara{The Hall of Supreme Harmony blazed with a thousand lanterns, their light catching the gold-leaf dragons coiled along the crimson pillars. Incense rose in perfect spirals from bronze tripods, and the air itself seemed to hold its breath as the Qin Emperor ascended the jade dais, his robes sweeping the marble steps like a river of black silk.}

\storypara{Wei Liang stood at his post among the honor guard, his spine locked into the perfect vertical demanded by the court style. His hands gripped his spear with the exact pressure prescribed—neither too tight nor too loose—and his gaze remained fixed on a point exactly three handbreadths above the Emperor's crown. He had performed this stance a thousand times. It required no thought.}

\storypara{But thought was precisely the problem.}

\storypara{Even now, with the Chief Inquisitor's voice droning through the ceremonial proclamations, Wei Liang felt the ghost's presence like a second heartbeat beneath his ribs. The dream-void training had changed something fundamental in his body. The Flowing River Style now lived in his muscles, in the spaces between his joints, in the way his breath moved through his lungs. He could feel the rigid court style pressing against him like a cage, and beneath it, the water waiting to move.}

\storypara{He tried to suppress it. He clenched his jaw and forced his shoulders into perfect stillness.}

\storypara{It did not work.}

\storypara{The Chief Inquisitor, Master Sheng, had not risen to his position by missing details. He was a man built of angles and edges—his face all sharp cheekbones and hard jaw, his body lean and corded like a drawn bowstring. His style was the Iron Mountain, the most absolute of the court forms, and he had personally overseen the execution of seventeen heretics who had dared to practice the old, soft ways.}

\storypara{As he turned to address the assembled court, his eyes swept across the honor guard with the precision of a scalpel.}

\storypara{They stopped on Wei Liang.}

\storypara{Master Sheng's head tilted, just slightly. His nostrils flared. And then, with a deliberate slowness that silenced the murmuring courtiers, he stepped down from his position beside the throne and walked directly toward the young guard.}

\storypara{"Hold," he said, and the single word carried the weight of an imperial decree.}

\storypara{The ceremony halted. Musicians lowered their instruments. The Emperor's eyes narrowed from his throne, but he did not intervene. Master Sheng had earned the right to interrupt.}

\storypara{Wei Liang's heart slammed against his ribs, but he kept his face still. He had been trained for this. He had been trained for everything except what was happening inside him.}

\storypara{Master Sheng circled him slowly, his footsteps echoing on the marble. He stopped directly in front of Wei Liang, close enough that Wei Liang could smell the sandalwood oil in the Inquisitor's robes.}

\storypara{"There is something foul in your aura, guard," Master Sheng said, his voice soft and dangerous. "Something that does not belong to the Iron Mountain. Something that moves."}

\storypara{He reached out and pressed two fingers against Wei Liang's chest, just below the collarbone. The touch was light, almost gentle, but Wei Liang felt it like a brand. The ghost inside him stirred, and for a fraction of a heartbeat, his energy responded—a ripple, a flow, a deflection that was not quite a movement.}

\storypara{Master Sheng's eyes widened. Then they hardened into chips of obsidian.}

\storypara{"The Flowing River Style," he breathed. "I have not felt its touch in twenty years. I burned the last heretic who practiced it, and I burned his manual with him." He stepped back, his voice rising to fill the hall. "This guard has been corrupted. He has learned the soft way, the way of the traitor, the way that defies the absolute loyalty of the Qin."}

\storypara{The crowd gasped. Courtiers drew back as if Wei Liang carried a plague. The other guards shifted their grips on their spears, uncertainty flickering across their faces.}

\storypara{Master Sheng drew his blade—a straight, heavy sword designed for the Iron Mountain's crushing strikes. "Demonstrate your loyalty," he said. "The ghost that has infected you—the spirit of the manual you have consumed—it is still inside you. Break its form with a single, absolute strike. Purge yourself of heresy, and I will recommend leniency."}

\storypara{Wei Liang's mouth went dry. He could feel the ghost coiled in his chest, waiting, trusting him. He knew what Master Sheng was asking. He had seen it done before. A loyal guard would turn his own energy inward, would sever the connection with a blow that would shatter the ghost and leave the practitioner scarred but alive.}

\storypara{But the ghost had taught him. The ghost had bled for him. The ghost had shown him his own face.}

\storypara{"No," Wei Liang said.}

\storypara{The word fell into the silence like a stone into still water.}

\storypara{Master Sheng's face went pale with rage. "You refuse?"}

\storypara{"I refuse."}

\storypara{The crowd erupted. Voices rose in shock and accusation. The Emperor's guards began to move, their boots thudding against the marble. Master Sheng's sword came up, and he lunged with the full weight of the Iron Mountain behind it—a strike meant to crush, to break, to end.}

\storypara{Wei Liang did not think. He could not think. The ghost moved through him, and he moved with it.}

\storypara{His body dropped, not in the rigid defensive crouch of the court style, but in a flowing spiral that carried him under the strike. His hand came up, palm open, and caught Master Sheng's wrist. Instead of blocking, he redirected—a gentle push, a turn of the hip, a shift of weight that seemed almost lazy.}

\storypara{Master Sheng's momentum carried him forward, past Wei Liang, directly into the jade pillar behind him. The impact cracked the stone. The Inquisitor crumpled, his sword clattering across the floor.}

\storypara{For a moment, there was absolute silence.}

\storypara{Then the guards charged.}

\storypara{They came from three directions, spears leveled, their faces masks of duty and fear. Wei Liang had nowhere to go. The Emperor had risen from his throne, his hand raised, his mouth opening to give the order that would mean death.}

\storypara{And then the ghost materialized.}

\storypara{It stepped out of Wei Liang's shadow, out of the air itself, out of the space between heartbeats. It was solid, scarred, its face a ruin of burn tissue that barely resembled human features. But Wei Liang knew those scars. He had seen them in his dreams. He had touched them with his own hands.}

\storypara{The ghost spread its arms, and the air rippled. The spear points struck an invisible wall, deflected as if they had hit water. The guards stumbled, off balance, their weapons skidding across the marble.}

\storypara{The ghost turned to Wei Liang. Its eyes held a weariness that mirrored his own dreams, a depth of sorrow that spoke of years lived and lost. It leaned close, and its voice was a whisper that cut through the chaos of the hall.}

\storypara{"Now you understand. I am you, from a future where you never burned the book. You were executed. I was sent back to ensure you make the choice that lets you live."}

\storypara{The words landed like a blade between Wei Liang's ribs.}

\storypara{He stared at his own scarred face, at the future that should have been his, and the weight of the choice crashed down upon him. The book. The fire. The ghost's first words, begging him not to burn it. The training in the dream-void. The scars that came from silence and knowing what was coming.}

\storypara{It was not a ghost. It was not a memory. It was him.}

\storypara{And he had been sent back to save himself by making himself burn the book.}

\storypara{The guards regrouped. Master Sheng rose, blood streaming from his forehead, his eyes burning with murder. The Emperor's voice cut through the hall, sharp as a blade: "Seize them both."}

\storypara{Wei Liang looked at the ghost—at himself—and saw the question in its eyes.}

\storypara{\emph{Now what?}}

\storyheading{Chapter 5: The True Choice}

\storypara{The burning courtyard from that first night surrounded them once more, though the flames had long since consumed the forbidden manual. Now torches ringed the space, held by two dozen imperial archers whose bows were drawn taut, arrowheads glinting like the eyes of hungry wolves. Captain Zhao stood at their head, his face carved from stone, while behind him, the Emperor's death warrant lay unfurled on a silk cushion carried by a trembling eunuch.}

\storypara{Wei Liang stood between the ghost and his doom.}

\storypara{The ghost's form flickered in the torchlight, half-solid, half-smoke, its face bearing the same scars that had first appeared in the dream-void. It did not look at the archers. It looked only at Wei Liang, and its eyes held a sorrow so deep it seemed older than the empire itself.}

\storypara{"You understand now," the ghost said, its voice carrying only to Wei Liang's ears.}

\storypara{And Wei Liang did understand. The terrible trap revealed itself like a blade turning in his chest. If he embraced the ghost—if he allowed himself to be consumed by this future self, to merge with the memory of what he would become—the archers would loose their arrows. He would die here, and history would reset. The ghost would find another "him" in another timeline, another young guard who burned the manual, another chance to break the chain.}

\storypara{But if he rejected the ghost—if he denied it, turned away, let it fade into the smoke from which it came—then the Flowing River Style would die forever. No one else would learn it. No one else would carry its whisper through the centuries. The art would become a forgotten dream, and the Qin court's rigid chains would bind every generation to come.}

\storypara{Two choices. Both were death. One of the body, one of the soul.}

\storypara{"Choose," Captain Zhao commanded, his voice flat and final. "The Emperor's seal waits for your answer. Embrace the demon, or renounce it. There is no middle path."}

\storypara{But the ghost smiled.}

\storypara{It was a strange smile, knowing and gentle, the smile of a man who had seen every possible ending and found one that the universe had overlooked.}

\storypara{"You don't have to be me," the ghost said.}

\storypara{Wei Liang's breath caught.}

\storypara{"Train the art in your heart alone," the ghost continued, its form growing fainter, as if the very act of speaking this truth was costing it substance. "Let no one see. Let the court believe I am destroyed. In public, be the perfect guard—rigid, obedient, unquestioning. Let Captain Zhao see the guard he expects. Let Master Sheng believe his strike purged the corruption."}

\storypara{The ghost raised its hand, and for a moment, it was not a ghost at all, but a man—a man who had walked this path before, who had learned to live in the spaces between what was seen and what was true.}

\storypara{"In secret, be the river."}

\storypara{Wei Liang's eyes widened. The archers shifted, uncertain, for they could not hear the ghost's words. They saw only a young guard frozen before a phantom, and a phantom that was slowly dissolving like morning mist.}

\storypara{"You will live long enough to pass it on," the ghost whispered, its voice now barely a breath. "Not as a manual, but as a whisper. Find one student. Just one. Teach them to flow. Let them teach another. The chain will not be broken—it will become a current, hidden beneath the ice of the court's law."}

\storypara{"But how..." Wei Liang's voice cracked. "How do I live with the lie?"}

\storypara{The ghost's smile deepened, and for the first time, Wei Liang saw not sadness in its eyes, but hope.}

\storypara{"You already do," it said. "Every guard lives a lie. The lie that obedience is strength. The lie that the Emperor's will is just. The lie that a man's heart can be forged into a weapon without breaking. You will simply live a truer lie—one that serves the future instead of the present."}

\storypara{The ghost's hand reached out, and Wei Liang took it. But this time, there was no merging, no absorption. The ghost's fingers were cool, insubstantial, like touching the memory of a hand.}

\storypara{"When I am gone," the ghost said, "they will see only a loyal guard who resisted temptation. They will burn the empty air and declare victory. And you will kneel, and bow, and thank the Emperor for his mercy."}

\storypara{"And then?" Wei Liang asked.}

\storypara{"And then you will live," the ghost said. "Truly live. Not as a hollow shell, but as a secret river carving a canyon one drop at a time."}

\storypara{The ghost began to fade, its edges blurring into the torchlight. But before it dissolved entirely, it leaned close, and Wei Liang felt a warmth that had nothing to do with fire.}

\storypara{"The scars I carry," the ghost said, "are from silence. But your scars will be from hope. They will not show on your skin. They will show in the students you never name, the techniques you never write, the legacy that will outlast the Qin dynasty itself."}

\storypara{And then the ghost was gone.}

\storypara{Wei Liang stood alone in the burning courtyard, surrounded by archers, facing the Emperor's death warrant. He felt the ghost's absence like a hollow in his chest, but also something new—a current, flowing where emptiness had been.}

\storypara{He turned to Captain Zhao. He knelt. He pressed his forehead to the cold stone.}

\storypara{"I have purged the demon," he said, his voice steady. "I am the Emperor's loyal servant."}

\storypara{Captain Zhao studied him for a long moment. The archers lowered their bows by a fraction. The eunuch with the death warrant hesitated, then rolled the scroll closed.}

\storypara{"The Emperor will be pleased," Captain Zhao said, but his eyes narrowed, searching for the lie.}

\storypara{He did not find it.}

\storypara{Because Wei Liang was not lying. He had purged the ghost. He had let it go. He had chosen to become something the court could not see, could not name, could not kill.}

\storypara{He would be the perfect guard.}

\storypara{And in the secret hours of the night, when the moon was high and the shadows were deep, he would be the river.}

\storypara{The archers withdrew. The torches were extinguished. The courtyard fell into darkness and silence.}

\storypara{And Wei Liang rose, brushed the ash from his knees, and walked back into the palace, carrying a secret that would outlive every emperor, every dynasty, every chain that men could forge.}

\storypara{In his heart, the Flowing River Style began to flow.}

\storypara{And somewhere, in the spaces between what was and what could be, a ghost that was once him smiled, and faded into the current of time, knowing that this time, the chain had not been broken.}

\storypara{It had simply learned to bend.}

\storyheading{Chapter 6: The Legacy of Silence}

\storypara{The moonlight painted the courtyard in silver and shadow, the same stones Wei Liang had crossed a thousand nights. Years had carved their passage into his face—deep lines from the corners of his eyes, a scar that ran from temple to jaw, knuckles swollen from decades of gripping a sword he no longer drew in anger. He was a ghost in plain sight now, the perfect guard, invisible in his loyalty.}

\storypara{The boy came stumbling through the archway, a practice blade clutched in trembling hands. He could not have been more than sixteen, his uniform still stiff with newness, his eyes wide with the terror of a first night patrol. He froze when he saw Wei Liang moving in the courtyard.}

\storypara{Wei Liang did not stop. He could not stop. The moon was full, and the Flowing River Style sang in his bones on such nights, demanding release after years of silence. His body became water—pouring, circling, yielding. A strike that was not a strike. A block that was an invitation. His feet traced characters on the flagstones, and his breath made the air shimmer.}

\storypara{The boy's practice blade clattered to the ground.}

\storypara{"What is that?" the apprentice whispered, his voice cracking with awe and fear. "I've never seen... that is not Qin combat. That is not anything I have been taught."}

\storypara{Wei Liang's movement ceased. He stood in the center of the courtyard, his shadow long and still, and he looked at the boy with eyes that had seen too much. The ghost had not visited him in years—not since the night he had knelt before the archers and declared it purged. But he felt its presence now, in the boy's unguarded wonder, in the question that could destroy them both.}

\storypara{To teach was to die. To teach was to undo thirty years of perfect silence, to invite the Inquisitors into his home, to paint a target on this boy's back before he had learned to sharpen his own blade. The court's chains were forged from such secrets. One word, one movement, one whispered lesson, and the Emperor's justice would fall like a hammer.}

\storypara{But to stay silent was to let the Flowing River Style die with him. The ghost had given him a third path—carry the art in his heart, pass it to a single hidden student. He had waited thirty years for a student. He had waited thirty years for a boy brave enough to ask the question.}

\storypara{Wei Liang crossed the courtyard slowly, his boots making no sound. He placed a hand on the boy's shoulder, feeling the tremor of youth, the hunger for something more than the rigid forms of the Qin court. The boy looked up at him, desperate for an answer.}

\storypara{"Forget what you saw," Wei Liang said.}

\storypara{The boy's face fell. He opened his mouth to protest, but Wei Liang's grip tightened, just once, a silent command. Then Wei Liang turned and walked away, his back straight, his duty fulfilled.}

\storypara{But as he passed the courtyard's fountain, his fingers brushed the water's surface. A single dried leaf, brown and brittle, floated there in the moonlight. He did not look back. He did not need to.}

\storypara{The boy stared at the leaf. It was shaped like a river's first bend, like the opening stroke of a character he had never been taught. He watched it spin in the fountain's gentle current, and something stirred in his chest—a question that would not be silenced, a hunger that would not be forgotten.}

\storypara{Wei Liang disappeared into the shadows of the palace corridor. Behind him, the leaf continued to float, unbroken and unseen, carrying the memory of water in a world of stone. And the boy, still staring, began to trace the shape with his finger in the air.}

\endgroup

\storyblockend

\storyblockstart
{Expanded StorySpark story under the \textit{Martial Arts} theme.}
{tab:case-storyspark-story}
{StorySpark story}

\begingroup
\footnotesize
\setlength{\parindent}{0pt}

\storyheading{Chapter 1: Ruined Imperial Conservatory}

\storypara{The air within the shattered dome of the Imperial Conservatory still carried the scent of rotted sandalwood and pulverized jade. Moonlight bled through a jagged fissure in the ceiling, illuminating the scattered bones of music stands and the skeletal remains of a hundred zithers. Meilin stood in the center of the desecrated hall, her breath misting in the cold, her fingers wrapped in blood-stained bandages.}

\storypara{Her guqin was a monstrosity now. The original lacquer had been stripped away, replaced with iron bands and a mechanism of her own design—a crank that could wind the seven strings to a razor edge. The seventh string, the one she called \emph{Wrath’s Echo}, could shear through a man’s throat at thirty paces.}

\storypara{Tonight, she would prove it.}

\storypara{From the shadows at the hall’s edge, a scout emerged. He wore the grey robes of a null-qi practitioner, his face blank, his hands raised in a defensive gesture. He was one of her student’s scouts, sent to test her.}

\storypara{Meilin’s fingers found the seventh string. She plucked.}

\storypara{The note was not a note. It was a scream of metal, a high-pitched \emph{shing} that split the air. The string detached from the bridge and shot forward, a silver blur aimed at the scout’s chest.}

\storypara{But the scout did not flinch. He raised one hand, palm open, and a silence-field expanded. The razor-string hit the shimmering barrier and \emph{stopped}. It hung in midair, quivering, as if caught in amber. Then it went slack. The field had neutralized the qi she had imbued into the steel, turning her weapon into nothing more than a limp piece of wire. It fell to the flagstones with a pathetic clatter.}

\storypara{Meilin’s heart seized. The scout took a step forward, and the silence-field crept closer. She could feel it now—a pressure against her eardrums, a deadening of sound. Her own heartbeat became a muffled thump, then a whisper, then nothing. The world was collapsing into a void of absence.}

\storypara{She stumbled backward, her sandals scraping against the rubble. Her hands, her broken hands, fumbled for the guqin’s body. She had no other weapon. She had staked everything on that one string.}

\storypara{And then she felt it.}

\storypara{A vibration. Not from her hands, not from her chest, but from the floor. A low, resonant hum that traveled up through the soles of her feet, into her spine, into the hollow of her skull. It was a note, ancient and pure, buried beneath centuries of stone and dust.}

\storypara{The scout paused. His null-qi field flickered.}

\storypara{Meilin looked down. Near the base of the ruined stage, a glint of tarnished bronze caught her eye. An old tuning fork, half-buried in the rubble, its prongs still quivering from the sympathetic resonance of her failed strike. The vibration grew stronger, and the entire conservatory seemed to groan in response. The walls, the pillars, the shattered dome—they all sang a single, mournful note.}

\storypara{She stared at the floor, at the intricate patterns of stone that had been laid centuries ago—not as decoration, but as a diagram. A harmonic ley-line nexus. The conservatory had not been built here by chance. It had been built to \emph{channel} something.}

\storypara{Her vulnerability was exposed, but so was the truth beneath her feet.}

\storyheading{Chapter 2: The Whisper-Gorge}

\storypara{The Whisper-Gorge was a wound in the earth that remembered.}

\storypara{Meilin had chosen this place for its acoustics, for the way the wind curled through the narrow passes and turned stone into song. She had mapped every resonance point in her youth, when she was still the Emperor's favorite musician and the gorge was her secret practice hall. Now she lay flat against the cliff's edge, the guqin strapped across her back, and watched General Kael's column march into the throat of the canyon below.}

\storypara{They moved in perfect silence.}

\storypara{Five hundred soldiers, their boots wrapped in felt, their armor padded with silk, their breath disciplined into nothingness. They carried no drums, no horns, no banners that might flutter and snap. Each man walked with the careful tread of a ghost, and where they passed, the land itself seemed to hold its breath. The Whisper-Gorge, famous for its singing stones and echoing walls, had fallen mute.}

\storypara{Meilin's fingers found the iron bands of her guqin. The seventh string—\emph{Wrath's Echo}—hummed against her palm.}

\storypara{She had spent three weeks preparing this ambush. The razor-strings were not new; she had been modifying them since the conservatory fell, threading monofilament wire through the silk cores, coating them in ground glass and crushed obsidian. But the \emph{cascade} was her invention—a technique that required her to pluck not individual strings but the spaces between them, creating a fan of cutting edges that spread like a peacock's tail of death.}

\storypara{Kael's soldiers reached the center of the gorge.}

\storypara{Meilin waited.}

\storypara{She counted their breaths, measured their steps, felt the weight of their null-qi fields pressing against the earth like stones dropped into still water. The ley-lines beneath the gorge were deep, ancient, sleeping. Kael's silence-troops had not yet learned to dampen them. That was her window.}

\storypara{She dropped.}

\storypara{Not straight down—that would have been suicide, a target against the sky. She fell in a spiral, her body twisting through the air, the guqin already drawn and positioned. The wind screamed past her ears, but she did not hear it. She heard only the strings.}

\storypara{The first pluck was a \emph{shatter}.}

\storypara{Five strings, struck simultaneously with the flat of her palm, sent a wave of razor-edged wire arcing downward. The strings did not fly; they \emph{extended}, unspooling from the guqin's body in a cascade of silver death. They caught the sunlight and threw it back in a thousand glittering fragments, disorienting the soldiers below.}

\storypara{The second pluck was a \emph{weave}.}

\storypara{Her fingers danced across the remaining strings, sending cross-cutting lines that intersected with the first wave, creating a net of slicing edges that descended upon the column like a fisherman's cast. The soldiers raised their shields, but the strings found the gaps—the spaces between armor plates, the exposed throats of men looking up, the hands gripping spear shafts too tightly.}

\storypara{Blood, and then sound.}

\storypara{The first scream broke the silence, and the gorge remembered how to speak. The walls caught the cry and multiplied it, throwing it back and forth until the canyon rang with the agony of Kael's soldiers. Their formation shattered. Men stumbled into each other, their null-qi fields flickering as concentration broke. The perfect discipline of silence collapsed into chaos.}

\storypara{Meilin landed on the shoulder of a fallen soldier, rolled, and came up with the guqin braced against her hip. Her fingers were already moving, finding the next pattern—a \emph{dirge} that would send vibration-shocks through the stone, disrupting their footing, turning the gorge floor into a drum of discord.}

\storypara{But Kael had not moved.}

\storypara{He stood at the center of the chaos, untouched by the blood and the screaming, his hands raised before him in a shape Meilin knew better than her own face. His fingers curled into the \emph{shang} mode—the second finger pressed to the thumb, the ring finger extended, the palm hollowed to catch and hold silence.}

\storypara{It was the same position she had taught him on the day he broke her hands.}

\storypara{"You taught me that, too," he said, and his voice carried through the chaos without effort, as if the silence itself bent to carry his words. "The \emph{shang} mode. The silence that answers a question with a question."}

\storypara{Meilin's fingers froze above the strings.}

\storypara{She watched, horror and recognition warring in her chest, as Kael's hands began to move. He was not playing a guqin—he had no instrument—but his fingers traced the same patterns, the same gestures, the same \emph{finger-positions} from her final lesson. And where his fingers moved, the silence followed.}

\storypara{The screams died first. Not abruptly, but as if they were being swallowed, each cry cut short by an invisible hand. Then the echoes faded. Then the wind itself fell still. Kael's null-qi field expanded, but it was no longer a mere suppression—it was a \emph{mirror}, a perfect antipathy that took every sound she had made and turned it back upon itself.}

\storypara{The razor-strings hung in the air, frozen mid-descent, quivering against the resistance of his silence-field. They had struck an invisible wall, and they could not advance.}

\storypara{Meilin's hands trembled. The broken bones in her fingers ached with the memory of that final lesson—the moment he had seized her wrists and \emph{twisted}, demanding she teach him the secret of the harmonic nexus. She had refused. And he had taken her refusal as permission to break her.}

\storypara{But now, as she watched his fingers curl into the familiar shape, she understood. He was not merely countering her attack. He was \emph{playing} silence against her, note for note, gesture for gesture, in perfect antipathy. Every move she had taught him, he now wielded as a weapon to undo her.}

\storypara{The gorge fell silent.}

\storypara{The ley-lines beneath the stone, which had begun to hum in response to her cascade, were smothered once more. Kael's silence-field pressed against her like a physical weight, and Meilin knew that her ambush had failed.}

\storypara{She had come to kill her student.}

\storypara{But he had already learned everything she had to teach.}

\storyheading{Chapter 3: Underground Resonance-Chamber}

\storypara{The stairwell spiraled downward, each step a descent into deeper darkness. Meilin pressed her palm against the damp stone wall, feeling the faint pulse of the ley-line nexus beneath the Conservatory. The tarnished tuning fork she had found in the ruined hall above still hummed in her pocket, its vibration a thread connecting her to something vast and ancient.}

\storypara{The air grew colder as she descended. The silence here was different from Kael's null-qi fields—it was not an absence of sound, but a \emph{waiting} for it. The stones themselves seemed to hold their breath.}

\storypara{At the bottom, a door of black iron stood half-open. Beyond it, a chamber carved from the living rock stretched into darkness. Meilin lit a small flame with her qi, and the light revealed what she had come to find.}

\storypara{The walls were covered in grooves.}

\storypara{Not writing, not symbols—but \emph{lines}. Thousands of them, carved into the stone at different depths and angles, some shallow as a fingernail's scratch, others deep enough to swallow a finger. They ran in spirals and curves, intersecting and diverging, forming patterns that seemed to shift when she looked at them directly.}

\storypara{She stepped closer, her breath misting in the cold air. The grooves were not random. They followed a logic—a grammar of absence that she recognized from years of teaching Kael the ancient forms of musical notation. But this was not music as she knew it. This was something else.}

\storypara{\emph{Tactile}, she realized. \emph{They were meant to be felt, not heard.}}

\storypara{Meilin knelt before the nearest wall and pressed her fingers into the first groove. It was long and slow, like a drawn-out sigh. She traced it, her fingertip following its path through the stone. The groove deepened, then shallowed, then deepened again—a rhythm of pressure and release.}

\storypara{She closed her eyes.}

\storypara{The groove was a silence. A specific silence, measured in the language of what it excluded. This one held the shape of a battlefield after the fighting had ended—the silence of bodies that would never move again, of cries that had been swallowed by the earth. She felt it in her bones: the weight of a thousand deaths, each one a note in this terrible symphony of absence.}

\storypara{\emph{This is his guilt}, she thought. \emph{This is what he carries.}}

\storypara{She moved to the next groove. It was shorter, sharper, like a cut. The silence of a door closing. The silence of a word swallowed before it could be spoken. The silence of a hand raised, then lowered.}

\storypara{Each groove was a different time-signature of silence. Some lasted centuries, their grooves winding around the entire chamber like the coils of a sleeping serpent. Others were mere heartbeats—brief, violent absences that ended in jagged edges where the silence had been broken.}

\storypara{Meilin traced them all. Her fingers grew raw, then numb, then raw again. She lost track of time. The chamber became her world, the grooves her language, and the silence her teacher.}

\storypara{She found the silence of the day she had first taken Kael as a student—a young boy with empty eyes and hands that trembled when he held a zither. The groove was hesitant, uncertain, like the silence between two strangers meeting for the first time.}

\storypara{She found the silence of the day he had played his first complete melody—a simple folk tune that had made her weep. The groove was warm, almost alive, filled with the silence of pride and promise.}

\storypara{She found the silence of the day he had broken her hands.}

\storypara{That groove was different. It was deeper than the others, carved with more force, as if the hand that made it had been shaking. It spiraled inward, tighter and tighter, until it ended in a point so sharp it drew blood from her fingertip. The silence of that moment was not empty—it was \emph{full}. Full of rage, yes, but also something else. Something that tasted like regret before it had learned its name.}

\storypara{\emph{He was crying}, she realized. \emph{When he did it, he was crying.}}

\storypara{Her blood dripped onto the stone, and the chamber hummed in response. The ley-line nexus beneath her feet vibrated, and the grooves on the walls seemed to pulse with a dim, inner light.}

\storypara{Meilin pulled her hand back and pressed it to her chest. Her heart was pounding. She had been tracing the melody for hours—or was it days?—and she was only now beginning to understand its true shape.}

\storypara{The lost melody was not a sequence of notes. It was a sequence of \emph{absences}—a map of every moment Kael had chosen silence over speech, every wound he had inflicted by withholding, every guilt he had buried in the hollow spaces of his memory.}

\storypara{She followed the grooves to the far wall, where the patterns converged into a single, final line. This was where the melody should end—where the final silence should complete the composition and bring it to resolution.}

\storypara{But the groove was incomplete.}

\storypara{It began like the others, a clean cut in the stone, but after a few inches it became ragged, uneven, as if the carver had lost their nerve. The line stuttered, then stopped altogether, leaving a raw, unfinished scar in the rock.}

\storypara{Meilin pressed her fingers to the broken edge. The stone was warm here, warmer than the rest of the chamber. She felt a faint residue of qi—not the vibrant, flowing qi of a musician, but something else. Something hollow and still.}

\storypara{\emph{Null-qi.}}

\storypara{She had felt it before, on every battlefield where Kael had silenced her strings. It was the absence of qi, the negation of vibration, the void that sound could not cross. And it was here, embedded in the stone like a fossilized whisper.}

\storypara{\emph{Kael had written this melody after crippling her.}}

\storypara{The realization struck her like a physical blow. She staggered back, her hand flying to her mouth. He had come here—to this hidden chamber beneath the Conservatory, to this place where the ley-lines hummed with the memory of ancient harmonies—and carved his guilt into the earth with his own null-qi.}

\storypara{The silence-fields he used against her were not just weapons. They were \emph{confessions}. Every time he silenced a battlefield, he was repeating a line from this melody. Every time he froze her strings, he was carving another groove into his own soul. Every time he looked at her with those empty eyes, he was reading the silence he had written into the stone.}

\storypara{But the final groove remained unwritten. The melody was incomplete. And until it was finished, neither of them could be free.}

\storypara{Meilin stood slowly, her fingers still tingling from the stone's cold story. She looked at her guqin—at the razor-edged strings, the iron bands, the seventh string she had named \emph{Wrath's Echo}. It felt hollow now, a child's tantrum against a storm. The question hung in the silence of the chamber, unanswered: \emph{If I finish the melody for him, will it end us—or begin something new?}}

\storypara{She turned back to the wall, to the incomplete groove that waited like an open wound. Her hand moved to her guqin, hovering over the strings. Somewhere above, in the ruined Conservatory, the tuning fork she had found still hummed with the resonance of the ley-lines. And somewhere beyond that, in the silence of the Whisper-Gorge, Kael was waiting.}

\storypara{The melody was not a weapon. It was a bridge. And the final silence was not an ending—it was an invitation.}

\storypara{Meilin closed her eyes and listened to the chamber's breath. The grooves on the walls seemed to whisper to her, each one a memory of pain and regret. She heard Kael's silence in every one of them—the silence of a boy who had been taught that sound was weakness, that music was a lie, that the only truth was the void.}

\storypara{But she had taught him differently. She had shown him that silence could be a foundation, not a prison. That absence could create space for something new.}

\storypara{And now, in this underground chamber, she understood what that something was.}

\storypara{She pressed her palm to the final, incomplete groove. The stone was warm against her skin, and she felt the null-qi residue pulse like a dying heartbeat. It was waiting for her. Waiting for the one who had been silenced to choose whether to continue the silence—or break it.}

\storypara{The ley-line nexus hummed beneath her feet, and the tuning fork in her pocket vibrated in answer. The chamber was alive with potential, a song waiting to be completed.}

\storypara{Meilin opened her eyes. Her reflection stared back at her from the polished surface of the seventh string—\emph{Wrath's Echo}, she had called it. But wrath was not what this moment required.}

\storypara{She lifted her hand from the stone and placed it on the strings of her guqin. The razor-edged wires hummed under her touch, eager and dangerous. But she did not pluck them. Instead, she waited.}

\storypara{The silence of the chamber deepened, and in that silence, she heard it: the faint, almost imperceptible vibration of the final groove, still incomplete, still waiting. It was not a demand. It was a question.}

\storypara{\emph{Will you finish what I started?}}

\storypara{Meilin's fingers hovered over the strings. The answer was not in the notes she could play, but in the silence she could choose to hold.}

\storypara{And in that moment of perfect stillness, she understood what the final groove required. Not music. Not sound. But a silence so complete, so intentional, that it could meet Kael's null-qi as an equal—and forge something new from the space between them.}

\storypara{She lowered her hands to her lap and closed her eyes.}

\storypara{The chamber held its breath.}

\storypara{And Meilin began to listen to the silence that was waiting to be written.}

\storyheading{Chapter 4: The Silent Dunes}

\storypara{The dunes stretched endlessly beneath a sky bleached of color, each grain of sand holding its breath as Kael descended from the ridge. Meilin had tracked him for three days across the Wasting Expanse, following the trail of silenced oases and birds that had fallen mid-song from the sky. Now he stood before her, no longer the boy she had taught, but something carved from the very absence she had only begun to understand.}

\storypara{"You shouldn't have come, Master." His voice was flat, devoid of the tremor she remembered. "The melody ends here."}

\storypara{Meilin's fingers hovered over \emph{Wrath's Echo}, the seventh string she had reforged with iron and rage. But before she could pluck it, Kael raised his hand—not in threat, but in offering. The air around them \emph{collapsed}.}

\storypara{It was not silence as she knew it. This was silence as a physical force, pressing against her eardrums until her own heartbeat became a phantom she could no longer hear. The wind died. The sand ceased its eternal whisper. Even the blood in her veins seemed to flow without sound, as if the universe had been muted at its source.}

\storypara{Her guqin trembled. She struck the seventh string—and felt nothing. No vibration traveled up her arm. No razor-edge sliced through the air. The string simply \emph{was}, and then it was not, its energy swallowed by the null-qi field that now surrounded them like a second skin.}

\storypara{She struck again. And again. Each time, the sound died before it could be born.}

\storypara{\emph{Weaponless.}}

\storypara{The word echoed in her mind, but even that thought felt muffled, as if her own consciousness was being wrapped in cotton. She had prepared for battle. She had prepared for death. She had not prepared for \emph{nothing}. Her guqin, her strings, her qi—all useless. She had only her instincts, raw and animal, screaming at her to move, to dodge, to survive.}

\storypara{Kael watched her, his eyes the color of ash. "Do you remember the first lesson you taught me? Before the Conservatory, before the politics, before I learned what silence could \emph{do}?"}

\storypara{Meilin's throat constricted. She remembered. A boy of seven, found wandering the streets with no voice and no past, drawn to her music like a moth to flame. She had taught him that silence was the canvas upon which sound painted its stories. She had never imagined he would learn to paint \emph{with} the canvas itself.}

\storypara{The sand beneath their feet began to shift.}

\storypara{At first, Meilin thought it was her failing vision, a trick of the null-qi pressing against her senses. But no—the grains were moving with purpose, carving patterns that spiraled outward from Kael's feet. Not random. Not chaotic. \emph{Familiar}.}

\storypara{She had traced those patterns with her fingers in the darkness of the underground chamber. She had memorized their contours, their gaps, their silences. The sand was writing the lost melody—not in sound, but in absence. Each groove that formed was a moment of silence from the ancient song, a wound carved into the earth as if the land itself remembered and wept.}

\storypara{The silence of the first note appeared: a jagged trench that split the dune in two. The silence of the second followed: a perfect circle, empty at its center. The silence of every pause in the forgotten melody: a constellation of hollow points, each one a star that had died before it could burn.}

\storypara{Meilin fell to her knees. Not from weakness, but because the sand was \emph{singing} to her in a language she had forgotten. The grooves were not random. They were the lost melody made visible, and the land itself was weeping as it remembered what sound had been stolen from it.}

\storypara{Kael's composure cracked. "Stop," he whispered, but no sound escaped his lips. She saw the word form, saw the pain twist his features, but the null-qi field swallowed his voice whole.}

\storypara{The sand continued its lament. A new pattern emerged, larger than all the others: a spiral that wound inward toward a single point at its center. Meilin recognized it from the chamber's final groove—the incomplete line, the one that ended in null-qi, the one that had been waiting for \emph{her}.}

\storypara{She looked up at Kael, and for the first time, she saw not the warlord who had broken her hands, but the boy who had cried while doing it. The boy who had carved his guilt into stone because he had no other way to confess.}

\storypara{But she had no weapon. No guqin. No qi. Only instinct.}

\storypara{And instinct told her to wait.}

\storypara{The sand patterns continued to form, each groove a silence from the lost melody, each hollow a wound in the earth's memory. The dunes wept in patterns, and Meilin watched, weaponless, as the land remembered what she could not play.}

\storyheading{Chapter 5: The Center of the Silent Dunes}

\storypara{The stillness between them had become a living thing, breathing with the weight of all that had been broken and all that remained unsaid. Meilin's fingers hovered over the single remaining string of her guqin—\emph{Wrath's Echo} lay shattered at her feet, its iron bands scattered across the sand like the bones of a forgotten war. Only this one string remained, still taut, still waiting.}

\storypara{Kael stood motionless before her, his null-qi field pressing against the air like a held breath. The sand between them had fallen still, the incomplete spiral of the lost melody carved into its surface now seeming to wait with her.}

\storypara{She did not think. She did not plan. She simply understood, in the way she had once understood how to make a blind boy feel music through his chest, that this moment required something beyond attack or defense.}

\storypara{Meilin plucked the string.}

\storypara{The note was pure, untainted by qi, unburdened by rage. It was not \emph{Wrath's Echo} that sang, but something older—the first lesson she had ever learned about sound: that vibration needed no ears to be felt.}

\storypara{The resonance traveled through the guqin's body, through the wood her hands had once known as home, and into the air between them. But it did not stop there. The vibration passed through Kael's sternum, through the cage of his ribs, through the marrow of his bones, bypassing his deaf ears entirely and reaching the core of him where no silence-field could follow.}

\storypara{Kael's null-qi core reacted before his mind could—an instinct honed through years of war, trained to dampen, to cancel, to silence. The vibration began to attenuate, to fade into the void he carried within him.}

\storypara{But the sensation triggered something.}

\storypara{A memory, unbidden and sharp: Meilin pressing his small hands to the belly of a guqin when he was seven, guiding his palms to feel the thrum of a single note through his chest because his ears could not hear it. \emph{This is how music lives in you,} she had said. \emph{Not through sound. Through feeling.}}

\storypara{His null-qi core hesitated.}

\storypara{The vibration did not stop.}

\storypara{Kael understood then what she was offering—not an attack, not a plea, but an invitation. He opened his mouth, and from the depths of his null-qi core, he hummed.}

\storypara{It was not a note. It was an anti-note, a pure null-frequency that should have canceled all sound. But he did not aim it at her vibration. He aimed it beside it, around it, a counterpoint of emptiness that wrapped around her resonance like a hand cupping a flame.}

\storypara{The two vibrations met.}

\storypara{They did not cancel.}

\storypara{They harmonized.}

\storypara{A third resonance emerged from their intersection—not sound, not silence, but something that existed in the space between. A soundless peace, born not from the absence of conflict but from the presence of two broken things choosing to vibrate together.}

\storypara{The stillness held them.}

\storypara{And in that stillness, the lost melody was complete.}

\storyheading{Chapter 6: The Entire Wounded Kingdom}

\storypara{The harmonic silence spread from Meilin and Kael like ripples in a pond that had forgotten water. It moved outward across the Wasting Expanse, where the sand that had carved itself into patterns of guilt and memory now smoothed into something like peace. The silence did not erase—it mended. Where null-qi scars had cracked the earth, the vibration threaded through like a needle pulling closed a wound. Where sound had been deadened for years, the air began to hum again, first at frequencies too low to hear, then in the whisper of wind across stone, the distant call of a bird testing its voice. The mending spread across the entire kingdom, each healed scar restoring sound in balanced measure—not too loud, not too soft, but exactly as it should be.}

\storypara{Meilin felt it in her chest first, a warmth that had nothing to do with qi and everything to do with the vibration still trembling between her and the boy she had once taught. Kael stood across from her, his eyes wide, his hands pressed against his own sternum as if feeling his heartbeat for the first time. The null-qi fields that had surrounded him like armor had dissolved into something softer, a silence that listened rather than consumed.}

\storypara{Neither of them spoke. Words felt unnecessary, almost clumsy, in the presence of what they had created.}

\storypara{The peace lasted three days.}

\storypara{On the fourth morning, dust rose on the horizon. Meilin recognized the gait of old horses, the creak of wooden carts that had traveled far. She rose from where she had been sitting beside Kael, teaching him with her fingers pressed to his throat how to feel the vibration of his own voice without needing to hear it.}

\storypara{The caravan that emerged from the haze was small but unmistakable. Survivors of the Imperial Conservatory—the ones who had hidden when Kael's purges came, the ones who had fled into the mountains and deserts, the ones who had watched their colleagues dragged away for refusing to silence their instruments. Meilin counted twelve faces she knew, each one etched with grief and hunger and a terrible, righteous fury.}

\storypara{At their head walked Master Wen, the oldest of the court musicians, his white beard stained yellow from years of hiding. He carried no instrument. His hands were empty, and that emptiness was itself a weapon.}

\storypara{"Meilin," he said, his voice cracking from disuse. "We felt it. The mending. We felt the land heal." His eyes moved past her to where Kael sat, still and watchful. "And we knew you had found him."}

\storypara{The others fanned out behind him, forming a semicircle that separated Meilin from her former student. They carried no blades, no bows, no weapons of steel. But their hands were shaped for strings and mallets and reeds, and in the old ways, those hands could kill just as surely.}

\storypara{"You know what must be done," Master Wen said. "He broke your hands. He silenced our halls. He murdered those who would not bow to his vision of perfect stillness." His voice trembled. "I buried my own daughter beneath the collapsed eastern wing. She was practicing the \emph{qin} when his soldiers came. She refused to stop playing."}

\storypara{Meilin closed her eyes. She had known. She had known about all of them.}

\storypara{"He is your student," another musician said, a woman whose left ear had been cut off as a warning. "Your failure to guide him. Your responsibility to end him."}

\storypara{The word hung in the air like a blade waiting to fall.}

\storypara{\emph{End him.}}

\storypara{Meilin opened her eyes and looked at Kael. He had not moved from his seated position, but his hands had come to rest on his knees, palms open. He was watching her with the same expression he had worn when she first taught him to feel music through his chest—trusting, terrified, hoping she would not abandon him.}

\storypara{"No," Meilin said.}

\storypara{Master Wen's face hardened. "You would spare the man who—"}

\storypara{"I would not spare him," Meilin interrupted, her voice quiet but carrying the weight of the harmonic silence still thrumming beneath their feet. "I would transform him. And I would transform you. And I would transform this entire wounded kingdom, not by cutting out what is broken, but by teaching it to sing in a new key."}

\storypara{She reached down and picked up a handful of sand from the expanse. Let it trickle through her fingers. Watched the grains catch the light.}

\storypara{"The lost melody," she said, "is not a sequence of notes. It is a sequence of silences. Kael taught me that, even when he did not mean to. He carved his guilt into stone, his pain into the very fabric of soundlessness. And when I finally listened—not to what he said, but to what he could not say—I found the melody waiting."}

\storypara{Master Wen shook his head. "You speak in riddles."}

\storypara{"I speak in truth." Meilin turned to face the survivors fully. "You want me to kill him because you believe justice requires blood. But I have already shed blood. I have already screamed. I have already felt my hands break beneath his, and I have already chosen to meet that breaking with something other than revenge."}

\storypara{She walked toward them, slowly, her feet bare against the healed earth.}


\storypara{"Join me." She began to hum—not a tune, but a silence shaped by breath. The first note of the lost melody, the one that said \emph{I remember}. She held it in her chest, let it resonate through her bones, and then let it fall away into stillness.}

\storypara{Master Wen's jaw tightened. But he did not look away.}

\storypara{Meilin hummed again. A different silence this time. The one that said \emph{I forgive}.}

\storypara{Behind her, she heard Kael's breath catch. She did not turn around. She kept her eyes on the survivors, on their grief-hardened faces, on the wounds they carried like instruments they no longer knew how to play.}

\storypara{One by one, they began to understand.}

\storypara{The woman with the missing ear was the first to join. She closed her eyes and let out a sound that was not quite a hum, not quite a sigh, but something in between—a silence that said \emph{I am still here}. The man beside her followed, his own breath shaping a pause that meant \emph{I remember her too}.}

\storypara{Master Wen was the last.}

\storypara{He stood rigid, his hands trembling at his sides, his face a mask of fury and sorrow and the unbearable weight of years spent running. Meilin watched him struggle. Watched him fight against the melody that was not a melody, the silence that was not empty.}

\storypara{And then he broke.}

\storypara{Not in anger. Not in collapse. He broke open, like a seed cracking in spring soil, and from his throat came a silence so raw and jagged that it tore the air itself—the silence of a father burying his daughter, the silence of a teacher watching his students die, the silence of a man who had forgotten that silence could be anything but violence.}

\storypara{The other survivors joined him, their collective silence rising like a chorus of held breaths, each one a different shade of loss and longing and the fragile possibility of something new.}

\storypara{Meilin stood at the center of them, her hands still scarred, her guqin shattered, her heart wide open. She felt Kael rise and walk toward her, felt his presence at her back, felt the null-qi in his core pulsing in time with her own vibration.}

\storypara{They were not enemies anymore. They were not teacher and student. They were co-creators of a soundless peace, born from mutual wounds, held together by the only melody that had ever mattered—the one that could only be played in the spaces between what was said and what was felt.}

\storypara{The harmonic silence spread further still, carrying their symphony of forgiveness across the kingdom, binding every scar and every silence and every broken thing into a new world that had learned, at last, how to listen.}

\endgroup

\storyblockend

\end{document}